\documentclass{article}
\usepackage{naturetex}

\usepackage{booktabs}
\usepackage{amsmath}  % For advanced math commands
\usepackage{amssymb}  % For \mathbb command
\usepackage{amsfonts}  % Alternatively, for \mathbb command
\usepackage{cuted}
\usepackage{makecell}
\usepackage{graphicx}
\usepackage{afterpage}
\usepackage[table,xcdraw]{xcolor}
\usepackage{lineno}
\usepackage{xurl}
\usepackage{nicematrix}
\usepackage{rotating}
\usepackage{adjustbox} % For resizing tables
\usepackage{multirow}  % For multi-row cells
\usepackage{caption}   % For customizing table captions
\usepackage{geometry}  % For page geometry
\usepackage{threeparttable}
\usepackage{siunitx}     % For formatting numbers
\usepackage{subcaption}

\newcolumntype{C}[1]{>{\centering\arraybackslash}p{#1}}

\usepackage{pdflscape}

\newlength{\hfoot}
\newlength{\vfoot}
\AddToHook {shipout/background} {\ifdim\textwidth=\linewidth\relax
\else\setlength{\hfoot}{-\topmargin}%
\addtolength{\hfoot}{-\headheight}%
\addtolength{\hfoot}{-\headsep}%
\addtolength{\hfoot}{-.6\linewidth}%
\ifodd\value{page}\setlength{\vfoot}{\oddsidemargin}%
\else\setlength{\vfoot}{\evensidemargin}\fi%
\addtolength{\vfoot}{\textheight}%
\addtolength{\vfoot}{\footskip}%
\raisebox{\hfoot}[0pt][0pt]{\rlap{\hspace{\vfoot}\rotatebox[origin=cB]{90}{\thepage}}}\fi}

\begin{document}
%\pagewiselinenumbers
% \linenumbers
\maketitle

{\setstretch{1.0}
	% *** ABSTRACT ***
\section*{Abstract}
Drug discovery is a time-consuming and expensive process, with traditional high-throughput and docking-based virtual screening hampered by low success rates and limited scalability. Recent advances in generative modeling, including autoregressive, diffusion, and flow-based approaches, have enabled \textit{de novo} ligand design beyond the limits of enumerative screening. Yet these models often suffer from inadequate generalization, limited interpretability, and an overemphasis on binding affinity at the expense of key pharmacological properties, thereby restricting their translational utility. Here we present Trio, a molecular generation framework integrating fragment-based molecular language modeling, reinforcement learning, and Monte Carlo tree search, for effective and interpretable closed-loop targeted molecular design. Through the three key components, Trio enables context-aware fragment assembly, enforces physicochemical and synthetic feasibility, and guides a balanced search between the exploration of novel chemotypes and the exploitation of promising intermediates within protein binding pockets. Experimental results show that Trio reliably achieves chemically valid and pharmacologically enhanced ligands, outperforming state-of-the-art approaches with improvements in binding affinity (+7.85\%), drug-likeness (+11.10\%) and synthetic accessibility (+12.05\%), while expanding molecular diversity more than fourfold. By combining generalization, plausibility, and interpretability, Trio establishes a closed-loop generative paradigm that redefines how chemical space can be navigated, offering a transformative foundation for the next era of AI-driven drug discovery.
}% setstretch

% *** INTRO ***
\section*{Introduction}
\label{sec1}
Drug discovery remains an exceedingly complex, costly, and time-intensive enterprise, typically requiring over a decade of sustained effort and substantial financial investment to translate a single therapeutic candidate into a clinically approved drug. Traditional high-throughput screening approaches have made important contributions, yet they are often constrained by low hit rates, escalating experimental costs, and limited coverage of the vast chemical space \cite{lyu2019ultra}. Docking-based virtual screening has provided a promising computational alternative, enabling the rapid prioritization of lead compounds and the identification of novel therapeutic opportunities. Nevertheless, these approaches remain hindered by high false-positive rates and intrinsic scalability bottlenecks, particularly as chemical libraries expand exponentially in both size and structural complexity \cite{li2024deep, zhou2024artificial}. Recent advances in generative modeling, however, represent a paradigm shift, offering a transformative capability to design novel lead compounds under task-specific optimization constraints. By directly generating molecular structures with desired properties, this emerging strategy not only mitigates the limitations of ultra-large library screening but also enables systematic exploration of previously inaccessible regions of chemical space \cite{jeon2025stella, cai2025fragment}.

Recent advances have introduced autoregressive generative models for designing ligands directly from protein 3D structural contexts. Representative approaches include Pocket2Mol, which leverages an E(3)-equivariant graph neural network to encode pocket geometry \cite{peng2022pocket2mol}, ResGen, which integrates pocket information with fragment-based autoregression \cite{zhang2023resgen}, and FragGen, which adopts fragment-wise generation guided by interaction graphs \cite{zhang2024fraggen}. While these models can condition molecular generation on protein features, their strictly sequential nature deviates from physical reality, accumulating errors that frequently yield chemically implausible structures \cite{du2024machine}. To overcome these issues, diffusion and flow-based models have emerged, offering simultaneous generation of all atoms and thereby capturing global interactions during the generative process \cite{guan20233d}. Notable examples include DiffBP, which employs diffusion denoising consistent with physical laws \cite{lin2025diffbp}, DiffSBDD, an SE(3)-equivariant conditional diffusion model \cite{schneuing2024structure}, and EquiFM, an equivariant flow-matching framework designed for greater efficiency \cite{song2023equivariant}. Collectively, these target-aware conditional generative models have shown promise in producing high-affinity ligands. Nevertheless, the limited availability of experimentally resolved protein–ligand complexes continues to impede model training, restricting their generalization and robustness in practical drug discovery applications \cite{krishnan2025generative}.

To overcome the generalization limitations of protein-conditioned generative models, researchers have increasingly drawn inspiration from language models, following the success of GPT in diverse domains. Molecular structures can be expressed in textual formats such as SMILES \cite{weininger1988smiles}, SELFIES \cite{krenn2020self}, and SAFE \cite{noutahi2024gotta}, enabling ultra-large compound libraries—once a bottleneck for virtual screening—to serve as valuable training corpora for molecular language models (MLMs). Representative efforts include BindGPT, which generates 3D ligands in protein binding sites through large-scale pretraining and reinforcement learning \cite{zholus2025bindgpt}; 3DSMILES-GPT, which augments SMILES with coordinate tokens for structure-aware generation \cite{wang20253dsmiles}; and TamGen, which incorporates protein embeddings via cross-attention and demonstrates efficacy in generating inhibitors against tuberculosis ClpP protease \cite{wu2024tamgen}. Despite improving generalization, current MLMs remain insufficient for precise protein-pocket targeting, and auxiliary optimization procedures often overemphasize binding affinity at the expense of drug-likeness (QED) and synthetic accessibility (SA), thereby limiting their translational utility in drug discovery \cite{sun2025synllama}.

\begin{figure}[htbp]
    \centering
    \includegraphics[width=\textwidth]{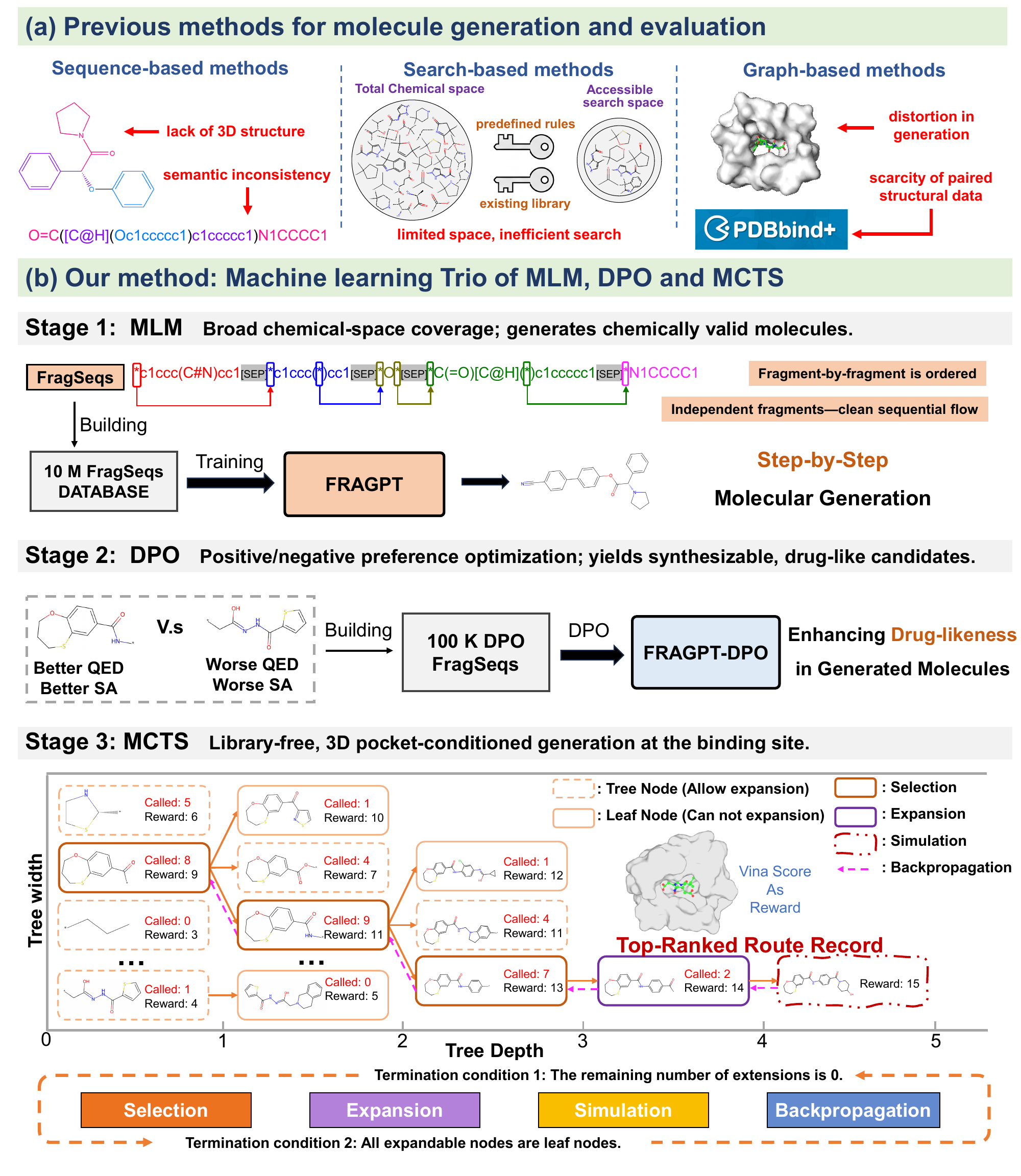}
    \caption{\textbf{Overview and motivation of the proposed Trio framework}. \textbf{a}, Limits of prior paradigms. Sequence-based (SMILES) models miss 3D context and inter-fragment semantics; search-based GA/MCTS depend on fixed fragment libraries and hand-crafted link rules, creating complicated and slow searches; structure-based 2D/3D generators need scarce protein-ligand pairs and risk geometric distortion. \textbf{b}, Trio pipeline. Stage 1: Pre-train: FRAGPT, a fragment language model trained on FragSeqs, learns context-aware attachments to assemble valid molecules step-by-step. Stage 2: Preference alignment, DPO with QED/SA pairs biases the policy toward synthesizable, drug-like compounds. Stage 3: Pocket-conditioned planning, the DPO-aligned policy drives MCTS with UCB over Selection-Expansion-Simulation-Backpropagation, combining affinity rewards to rank routes.}
    \label{fig:overall}
\end{figure}

In summary, while recent molecular generation models provide powerful means to navigate chemical space and design novel compounds, they often reduce molecular design to over-atomized or over-symbolized representations (Fig.~\ref{fig:overall}a). By prioritizing local interactions with binding-site residues through physics-based optimization, these approaches frequently neglect the semantic integrity of molecular functionality, undermining the plausibility of whole-molecule affinity \cite{zhang2025artificial}. Moreover, the limited interpretability of current models remains a fundamental barrier: their black-box nature obscures the pathways of molecular optimization, leaving chemists unable to rationalize or trust the design outcomes, thereby constraining their broader adoption in drug discovery \cite{zitnik2025ai}.

To jointly address generalization, plausibility, and interpretability, we propose Trio, a closed-loop paradigm that integrates a fragment-based MLM, reinforcement learning (RL), and Monte Carlo tree search (MCTS). At its core, a fragment-based MLM is trained on millions of SMILES strings to capture broad fragment sequence distributions and generate context-aware assemblies while circumventing the syntactic complexity of numeric junction identifiers and ring-index markers in SAFEGPT. To ensure drug-like plausibility, RL aligns the generative process with critical molecular properties such as QED and SA scores. Finally, the RL-aligned MLM then acts as the policy within MCTS, which explores fragment assembly trajectories in protein pockets using an upper confidence bound strategy to balance exploitation of promising structures against exploration of novel chemotypes, guided by affinity, pharmacokinetic, and SAR rewards. By combining fragment-level semantics, property-constrained optimization, and tree-based search, Trio achieves an interpretable and efficient molecular generation process that overcomes key limitations of prior approaches. Building upon this design, the backbone fragment-based MLM first demonstrates strong validity and novelty across both general \textit{de novo} and constrained generation tasks. For the target-based molecular generation setting, Trio establishes a new performance benchmark, significantly outperforming state-of-the-art approaches through a superior balance of physicochemical properties. The model achieves robust gains over current baselines, improving drug-likeness by 11.10\% and synthetic accessibility by 12.05\%. Crucially, these enhancements are not achieved at the expense of potency; rather, Trio concurrently elevates predicted binding affinity (+7.85\%) while notably expanding molecular diversity by fourfold. These results underscore the complementary strengths of fragment-informed MLMs, RL-driven property alignment, and MCTS-based strategic exploration, offering an effective and interpretable paradigm for targeted molecular design.

% *** RESULTS ***
\section*{Results}
\label{results}
\subsection*{Overview of Trio framework}
Trio can produce desired molecules from scratch for each target protein. Its overall generation procedures can be divided into three parts: first, using self-supervised learning to train an MLM for next fragment prediction tasks; second, adopting reinforcement learning to fine-tune the MLM for customized molecular property alignment; third, leveraging the Monte Carlo Tree Search and the aligned MLM to stepwise generate molecules in three-dimensional protein pockets.

The supervised MLM of Trio uses a GPT-like architecture, named FRAGPT, to predict molecule fragments in an autoregressive manner. Original SMILES strings of molecules need to be modified into fragment-based SMILES tokens for training. The fragmentation approach not only preserves intrinsic intra-fragment semantic information but also explicitly captures the chemical interactions between fragments. Subsequently, our supervised MLM employs a causal attention mechanism to generate molecular fragments step by step based on their contextual semantic environment, as shown in Fig.~\ref{fig:overall}b. Such a fragment-based generation strategy can effectively leverage the strong semantic feature extraction capability of MLMs and significantly reduce complexity compared to generating entire molecules directly.

However, the supervised MLM architecture alone is insufficient for generating molecules with desirable target properties, since GPT-like models inherently align with learned semantic distributions of tokens rather than explicitly optimized attributes. To address this limitation, the direct preference optimization (DPO) is used to fine-tune the supervised MLM by explicitly aligning the model's conditional distribution with an external preference signal that reflects the desired molecular attributes \cite{dpo}. The strategy can incorporate property preferences into the supervised MLM, similar to the human preference alignment training for large language models (LLMs) \cite{kopf2023openassistant}. Such an aligned MLM enables the production of druggable molecules simultaneously satisfying multiple targeted properties.

Furthermore, Trio combines the aligned MLM with an MCTS algorithm in the complicated target-aware molecular design. The hybrid approach leverages MCTS's strengths in balancing exploration and exploitation, facilitating a more diverse generation of molecules with enhanced binding affinities. The flexibility is an additional advantage of this paradigm, which allows straightforward adjustment of the search objectives through altering reward functions, circumventing the computational overhead associated with repeated fine-tuning. Moreover, this fragment-by-fragment search significantly enhances interpretability compared to fine-tuning approaches, because the optimization trajectory of molecular fragments transparently reflects the strategic decision-making process whereas fine-tuning interpretability remains constrained by black-box neural network weights.

\begin{figure}[htbp]
\centering
\includegraphics[width=\textwidth]{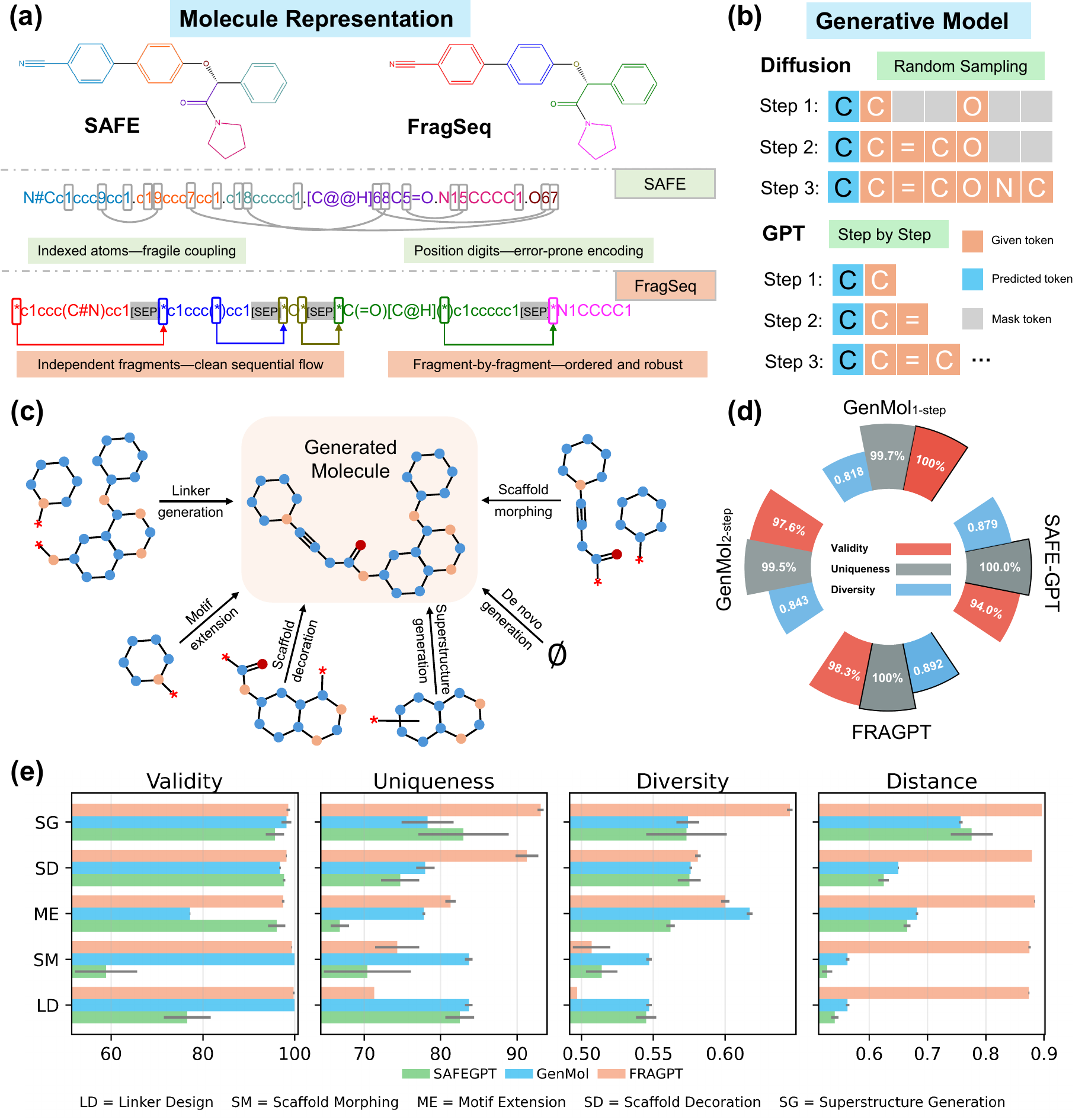}
\caption{\textbf{FRAGPT for \textit{De Novo} and Fragment-Constrained Molecular Generation: Representations, Models, Tasks, and Performance} \textbf{a}, Two fragment-based SMILES representations: SAFE and FragSeq, illustrating tokenization and ordering; \textbf{b}, Two language-model families for molecule generation: diffusion with random sampling and GPT with step-by-step masked prediction; \textbf{c}, Task taxonomy. Linker generation and scaffold morphing share the same conditional form but use different given fragments. Motif extension, scaffold decoration, and superstructure generation also share a common form, conditioned respectively on a motif, a scaffold, or a superstructure; \textbf{d}, \textit{De novo} generation: four models compared on the core metrics; \textbf{e}, Task-wise performance of three models across LD (Linker Design), SM (Scaffold Morphing), ME (Motif Extension), SD (Scaffold Decoration), and SG (Superstructure Generation) on Validity, Uniqueness, Diversity, and Distance. Validity is the percentage of chemically valid molecules. Uniqueness is the proportion of unique molecules among the valid ones. Diversity measures internal structural dissimilarity within the generated set. Distance measures structural similarity to a reference molecule; values approaching 1 indicate greater dissimilarity.}
\label{fig:fragpt}
\end{figure}

\subsection*{\textit{De novo} and Fragment-Constrained Molecular Generation}
To evaluate the performance of FRAGPT on molecular generation tasks, we conducted a comparative analysis against two existing fragment-based models: SAFEGPT \cite{safegpt} and GenMol \cite{genmol}. As illustrated in Fig.~\ref{fig:fragpt}a, a primary distinction lies in the encoding strategy: FRAGPT utilizes the FragSeq representation for molecular fragmentation, whereas both SAFEGPT and GenMol employ SAFE strings.
Furthermore, the baseline models differ in their underlying architectures. Specifically, SAFEGPT is built upon the GPT framework, while GenMol leverages a diffusion language model (Fig.~\ref{fig:fragpt}b). The evaluation framework focuses on two key aspects of generative chemistry: the capacity for creating novel chemical entities (\textit{de novo} generation) and the precision of fragment-based structural assembly (fragment-constrained generation), with the specifics of each task outlined in Fig.~\ref{fig:fragpt}c.
The quality of the generated molecules is assessed through three complementary metrics:
Validity, which quantifies the proportion of syntactically correct SMILES;
Uniqueness, which measures the percentage of non-redundant molecules among the valid ones; and Diversity, calculated as the average pairwise Tanimoto distance between the Morgan fingerprints of the generated molecules.
For fragment-constrained tasks, we introduce an additional metric, Distance, defined as the average Tanimoto distance between each generated structure and its corresponding reference molecule. This metric captures the degree of chemical exploration under fixed structural constraints.

As shown in Fig.~\ref{fig:fragpt}d, FRAGPT trained on merely 1\% of the SAFE dataset achieves or even surpasses the performance of baseline models trained on the full corpus in the \textit{de novo} generation task, demonstrating its remarkable data efficiency. The comparatively low validity of SAFEGPT arises from its reliance on positional numeric markers for fragment linkage. It scales poorly because these digits interfere with canonical ring-closure notation and elevate syntactic ambiguity with increasing fragment counts. FRAGPT avoids this failure mode by imposing a structured fragment syntax that disentangles junction semantics from ring indices, yielding both higher validity and greater structural diversity. Although the diffusion-based GenMol attains substantial validity, its conservative denoising schedule suppresses exploration, leading to a diversity deficit.

While FRAGPT demonstrates exceptional capability in \textit{de novo} molecular generation, its true strength lies in addressing the critical need for fragment-constrained molecular design—a cornerstone of lead optimization in drug discovery. We rigorously evaluate FRAGPT across five key tasks: scaffold decoration, scaffold morphing, linker generation, motif extension, and superstructure generation. Notably, as a left-to-right autoregressive model, FRAGPT inherently faces architectural constraints in tasks requiring simultaneous satisfaction of both start- and end-fragment conditions (linker design and scaffold morphing). To overcome this challenge, we implement a novel beam search strategy that systematically explores the chemical space while maintaining fragment constraints, achieving remarkable success in these demanding scenarios.

Fig.~\ref{fig:fragpt}e presents the results of fragment-constrained generation from 100 generated samples per task. Compared to the other methods, FRAGPT demonstrates consistent and superior performance across multiple critical metrics. Specifically, FRAGPT attains near-perfect validity in every task and delivers the highest structural-distance scores across the board, indicating a consistently broader exploration of chemical space than SAFEGPT or GenMol. Even within the structurally confined tasks of linker design and scaffold morphing, FRAGPT demonstrates remarkable generative diversity. Its resulting inter-molecule distance significantly surpasses all competing methods, indicating that its generated candidates populate more distant and novel regions of the chemical space. In less constrained generation settings such as motif extension, scaffold decoration, and superstructure elaboration, the model consistently achieves high uniqueness, broad exploration radius, and strong chemical fidelity, demonstrating both flexibility and precision.

\begin{figure}[htbp]
\centering
\includegraphics[width=0.97\textwidth]{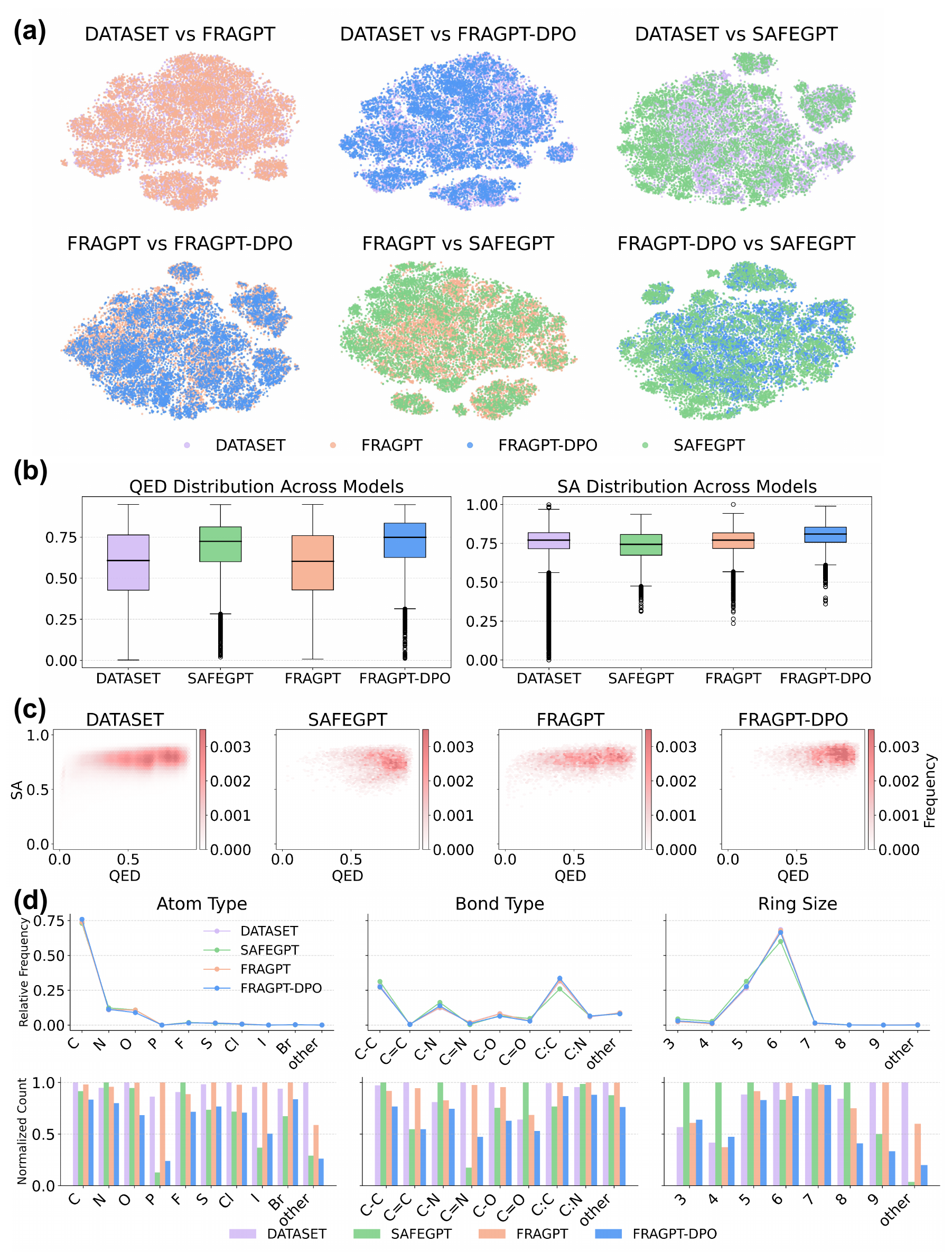}
\caption{\textbf{Comparative characterization of generated chemical spaces across baseline data and generative models.} \textbf{a}, Two-dimensional t-SNE of MACCS fingerprints of 10,000 generated molecules per set, showing pairwise overlaps between DATASET, FRAGPT, SAFEGPT and FRAGPT-DPO; \textbf{b}, Box plots of drug-likeness (QED) and synthetic accessibility (SA) for the same sets; \textbf{c}, Hexbin density maps of the QED-SA landscape; \textbf{d}, Statistical analysis of generated molecular substructures. A comparison of atom, bond, and ring distributions between the reference dataset and molecules from three generative models. (Top) Relative frequency plots show the proportion of each substructure category within each data source. (Bottom) Normalized count plots compare the prevalence of each substructure across the different sources, with values for each category scaled by the maximum observed count.}
\label{fig:rl}
\end{figure}

\subsection*{Drug-likeness Property Alignment}
\label{RL}
Building upon the robust generative capabilities demonstrated in the \textit{de novo} and fragment-constrained tasks, we employ the DPO algorithm to align our FRAGPT model with the drug-like scores in preparation for subsequent target-specific molecule-generation tasks. Here, the drug-likeness is quantified by the QED and SA metrics. To characterize the explored chemical space, we sampled 10,000 molecules from the training set and independently generated 10,000 molecules with SAFEGPT, vanilla FRAGPT, and DPO-aligned FRAGPT. In this experiment, MACCS fingerprints are embedded into a two-dimensional manifold via t-distributed stochastic neighbour embedding (t-SNE).

As illustrated in Fig.~\ref{fig:rl}a, vanilla FRAGPT almost completely spans the data manifold of the training set. The DPO-aligned variant (termed FRAGPT-DPO) further concentrates this distribution toward the data-dense core, whereas SAFEGPT preserves the central cloud but generates several additional high-density clusters that are sparsely represented in the original dataset, likely owing to its larger and more diverse training corpus. FRAGPT-DPO tends to compress the existing distribution and shift sample density inward, while SAFEGPT produces several new high-density clusters absent from the FRAGPT-DPO landscape.

To verify whether the contraction of FRAGPT-DPO's output distribution would compromise fundamental chemical realism, we compared the drug-like property distributions (QED and SA) across the aforementioned statistics. For the distribution of QED and SA, Figs.~\ref{fig:rl}b and~\ref{fig:rl}c reveal that vanilla FRAGPT closely mirrors the joint QED-SA landscape of the training set. Meanwhile, SAFE attains an improvement in the QED compared with vanilla FRAGPT, yet its SA distribution remains broader, suggesting a bias toward drug-likeness over synthetic accessibility. After DPO alignment, FRAGPT-DPO shows a clear upward shift in QED and a moderate improvement in SA, accompanied by a contraction in SA variance. The hex-bin plot reveals a marked shift in sample density toward the chemically desirable region, effectively eliminating the low-quality long tail present in the original data.

Besides basic drug properties, we also compared the relative frequencies of fundamental structural descriptors generated by three language-based molecular models with those in the training data. As shown in Fig.~\ref{fig:rl}d, the top panel demonstrates that all three generators closely reproduce the training-set statistics for atom-type, bond-type, and ring-size distributions. The bottom panel further reveals that vanilla FRAGPT preserves similar frequencies of all three descriptors, including low-frequency halogens (I, Br, Cl) and macrocycles. While this feature expands structural diversity, it also causes a decline in the SA and QED scores of the generated molecules. In contrast, SAFE retains too many small and large ring structures, resulting in inferior QED and SA compared to FRAGPT-DPO. Notably, FRAGPT-DPO abandons chemically unfavourable motifs, thereby improving drug-likeness and synthetic accessibility relative to the dataset.

\begin{figure}[htbp]
\centering
\includegraphics[width=0.97\textwidth]{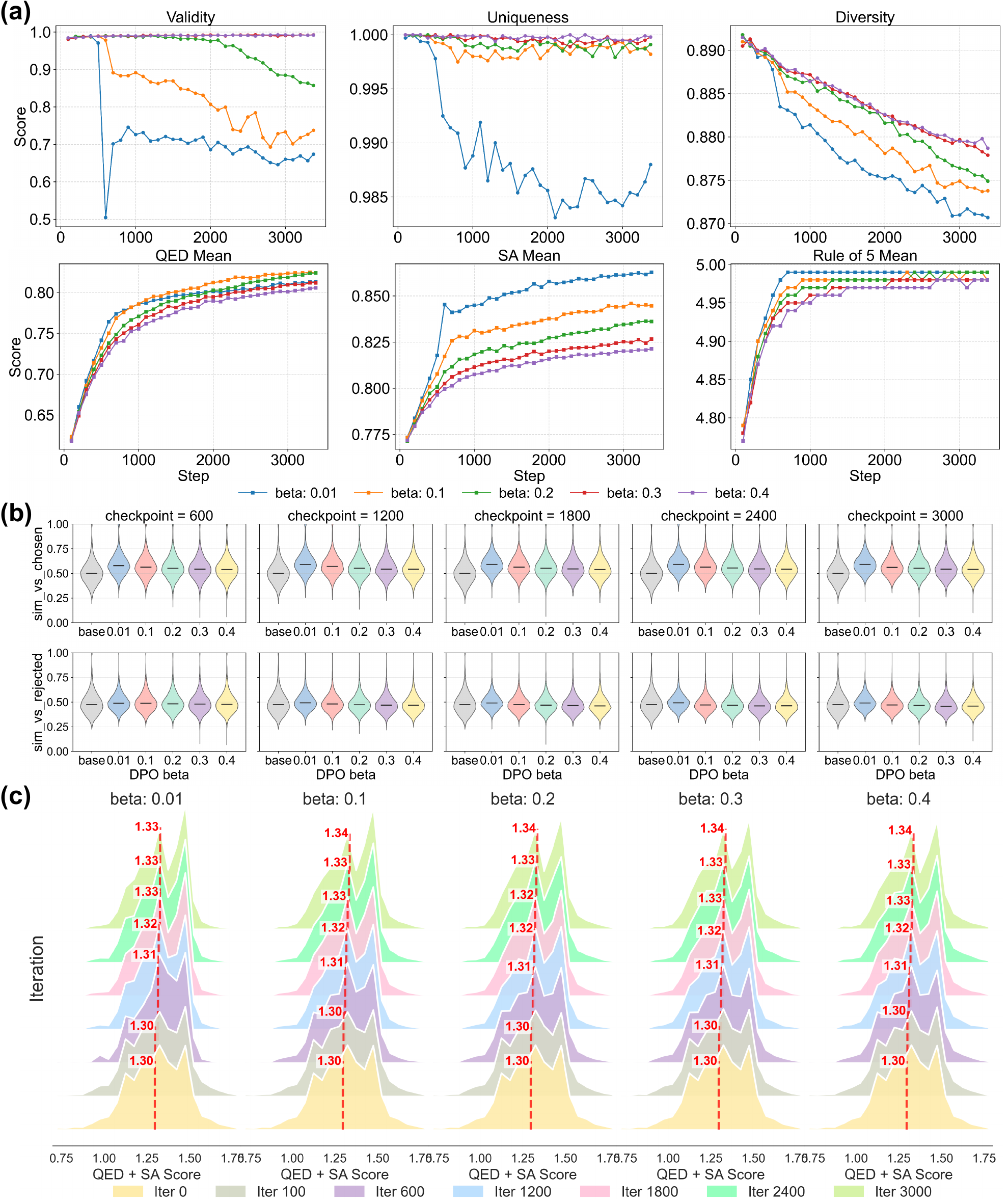}
\caption{\textbf{Impact of Direct Preference Optimization (DPO) regularization weight ($\beta$) on fragment-based molecular generation.} \textbf{a}, Trajectories of generative quality metrics (Validity, Uniqueness, Diversity) and target drug-likeness properties (QED Mean, SA Mean, Rule of 5 Mean) over 3,000 training steps. Line colors denote varying KL penalty weights, $\beta$, ranging from 0.01 to 0.4. The data illustrates the trade-off between rapid reward optimization under weak constraints and distribution preservation under strict penalties; \textbf{b}, Violin plots detailing the structural similarity of generated molecules to the chosen and rejected samples from the training dataset across successive checkpoints. Distributions remain concentrated around 0.50, indicating effective generalization without exact structural memorization; \textbf{c}, Ridgeline density plots showing the progression of combined QED and SA scores (QED + SA Score) for novel, low-frequency fragments across consecutive training iterations under different $\beta$ configurations. The progressive rightward shift demonstrates the model's sustained capability to construct chemically favorable novel substructures.}
\label{fig:dpo}
\end{figure}

\subsection*{Impact of KL regularization on generation dynamics and fragment-level novelty}
To steer our autoregressive transformer-based fragment generation model toward regions of chemical space with optimized drug-like properties, we applied Direct Preference Optimization (DPO). A critical challenge in aligning generative models via DPO is managing the trade-off between reward optimization and the divergence from the reference distribution. This balance is governed by the Kullback-Leibler (KL) regularization weight $\beta$. We systematically evaluated the impact of different values of $\beta$ on generation dynamics, molecule-level properties, and fragment-level novelty.

We first monitored the generation trajectories by sampling 10,000 molecules at 100-step intervals for each $\beta$ configuration, spanning from training step 100 to 3,400 (Fig.~\ref{fig:dpo}a). The model's capacity to improve target properties (specifically mean QED, mean SA, and mean Rule of 5 compliance) is highly sensitive to the $\beta$ constraint. Under weak regularization (e.g., $\beta=0.01$), the model strongly maximizes the reward, achieving the most rapid and highest gains in QED, SA, and Lipinski's Rule of 5 scores. However, this comes at the cost of reward hacking, evidenced by a substantial deterioration in standard generative quality metrics. Specifically, validity, uniqueness, and diversity experience a sharp decline as training progresses under $\beta=0.01$. Conversely, stringent KL penalties (e.g., $\beta=0.4$) strictly constrain the model to the reference distribution. While this effectively preserves validity near 1.0 and maintains structural diversity, it substantially decelerates the optimization trajectory, yielding the lowest improvements in QED and SA scores among all tested configurations. Intermediate constraints (e.g., $\beta=0.1$ or $0.2$) effectively balance this trade-off, significantly enhancing drug-likeness profiles while averting the severe mode collapse observed at lower $\beta$ values.

We further investigated whether the observed property gains are merely an artifact of the model memorizing highly rewarded molecules from the DPO dataset. Fig.~\ref{fig:dpo}b shows the structural similarity of the generated compounds to the positive and negative samples from the training data across various checkpoints. Regardless of the applied $\beta$ value, the similarity distributions are concentrated around 0.50 and exhibit no significant skewness toward 1.0. This confirms that the DPO-fine-tuned model generalizes effectively; rather than replicating the exact structures of the preferred dataset, it successfully captures the underlying chemical rules.

To ensure the model generates high-quality novel substructures rather than exploiting combinatorial shortcuts via high-frequency fragments, we analyzed the fragment-level quality distributions (Fig.~\ref{fig:dpo}c). By explicitly excluding high-frequency fragments found in the baseline dataset from this evaluation, we isolated the model's performance on novel sequences. The ridgeline plots demonstrate a consistent and progressive rightward shift in the density distributions of the combined QED and SA scores over successive iterations, observed across all $\beta$ settings. This sustained shift in density peaks confirms that the DPO objective guides the autoregressive model to construct chemically favorable, low-frequency fragments, thereby driving molecule-level improvements without relying on trivial structural repetition. Further detailed analyses regarding these fragment-level distributions are provided in the Supplementary Information.

\begin{landscape}
\pagestyle{empty}

\begin{table}[!htb]
\centering
\caption{Quantitative comparison of docking performance on five protein targets.}
\renewcommand{\arraystretch}{1.4} % Increase row height

\begin{threeparttable}
\begin{tabular}{|c|c|c|c|c|c|c|c|c|c|c|c|}
\hline
% Header row 1: Main Categories
\multirow{3}{*}{\textbf{Group}} & \multirow{3}{*}{\textbf{Method}} & \multicolumn{10}{c|}{\textbf{Target Protein}} \\ \cline{3-12}
% Header row 2: Specific Targets
& & \multicolumn{2}{c|}{\textbf{PARP1 $\uparrow$}} & \multicolumn{2}{c|}{\textbf{FA7 $\uparrow$}} & \multicolumn{2}{c|}{\textbf{5HT1B $\uparrow$}} & \multicolumn{2}{c|}{\textbf{BRAF $\uparrow$}} & \multicolumn{2}{c|}{\textbf{JAK2 $\uparrow$}} \\ \cline{3-12}
% Header row 3: Metrics (Mean/Std)
& & Mean & Std & Mean & Std & Mean & Std & Mean & Std & Mean & Std \\
\hline
\multirow{14}{*}{Baselines}
& JT-VAE \cite{jtvae} & 9.482 & 0.132 & 7.683 & 0.048 & 9.382 & 0.332 & 9.079 & 0.069 & 8.885 & 0.026 \\ \cline{2-12}
& REINVENT \cite{reinvent} & 8.702 & 0.523 & 7.205 & 0.264 & 8.770 & 0.316 & 8.392 & 0.400 & 8.165 & 0.277 \\ \cline{2-12}
& Graph GA \cite{graphga} & 10.949 & 0.532 & 7.365 & 0.326 & 10.422 & 0.670 & 10.789 & 0.341 & 10.167 & 0.576 \\ \cline{2-12}
& MORLD \cite{morld} & 7.532 & 0.260 & 6.263 & 0.165 & 7.869 & 0.650 & 8.040 & 0.337 & 7.816 & 0.133 \\ \cline{2-12}
& HierVAE \cite{hiervae} & 9.487 & 0.278 & 6.812 & 0.274 & 8.081 & 0.252 & 8.978 & 0.525 & 8.285 & 0.370 \\ \cline{2-12}
& GA+D \cite{g+d} & 8.365 & 0.201 & 6.539 & 0.297 & 8.567 & 0.177 & 9.371 & 0.728 & 8.610 & 0.104 \\ \cline{2-12}
& MARS \cite{mars} & 9.716 & 0.082 & 7.839 & 0.018 & 9.804 & 0.073 & 9.569 & 0.078 & 9.150 & 0.114 \\ \cline{2-12}
& GEGL \cite{gegl} & 9.329 & 0.170 & 7.470 & 0.013 & 9.086 & 0.067 & 9.073 & 0.047 & 8.601 & 0.038 \\ \cline{2-12}
& RationaleRL \cite{rationale} & 10.663 & 0.086 & 8.129 & 0.048 & 9.005 & 0.155 & \multicolumn{2}{c|}{No hit found} & 9.398 & 0.076 \\ \cline{2-12}
& FREED \cite{freed} & 10.579 & 0.104 & 8.378 & 0.044 & 10.714 & 0.183 & 10.561 & 0.080 & 9.735 & 0.022 \\ \cline{2-12}
& PS-VAE \cite{pavae} & 9.978 & 0.091 & 8.028 & 0.050 & 9.887 & 0.115 & 9.637 & 0.049 & 9.464 & 0.129 \\ \cline{2-12}
& MOOD \cite{mood} & 10.865 & 0.113 & 8.160 & 0.071 & 11.145 & 0.042 & 11.063 & 0.034 & 10.147 & 0.060 \\ \cline{2-12}
& RetMol \cite{retmol} & 8.590 & 0.475 & 5.448 & 0.688 & 6.980 & 0.740 & 8.811 & 0.574 & 7.133 & 0.242 \\ \cline{2-12}
& Genetic GFN \cite{gflow} & 9.227 & 0.644 & 7.288 & 0.433 & 8.973 & 0.804 & 8.719 & 0.190 & 8.539 & 0.592 \\
\hline
\multirow{2}{*}{\shortstack{Recent\\SOTA}}
& GEAM \cite{geam} & 12.891 & 0.158 & 9.890 & 0.116 & 12.374 & 0.036 & 12.342 & 0.095 & 11.816 & 0.067 \\ \cline{2-12}
& f-RAG \cite{f-rag} & $\mathit{12.945}$ & $\mathit{0.053}$ & 9.899 & 0.205 & $\mathit{12.670}$ & $\mathit{0.144}$ & $\mathit{12.390}$ & $\mathit{0.046}$ & $\mathit{11.842}$ & $\mathit{0.316}$ \\
\hline
\multirow{2}{*}{Ours}
& \textbf{Trio*} & $\mathbf{13.129}$ & $\mathbf{0.049}$ & $\mathbf{10.359}$ & $\mathbf{0.029}$ & $\mathbf{12.954}$ & $\mathbf{0.020}$ & $\mathbf{12.591}$ & $\mathbf{0.034}$ & $\mathbf{11.855}$ & $\mathbf{0.027}$ \\ \cline{2-12}
& \textbf{Trio} & 12.730 & 0.012 & $\mathit{10.132}$ & $\mathit{0.015}$ & $\mathit{12.669}$ & $\mathit{0.039}$ & $\mathit{12.389}$ & $\mathit{0.009}$ & 11.806 & 0.018 \\
\hline
\end{tabular}
\begin{tablenotes}
\footnotesize
\item Performance is evaluated using the mean Vina score of the generated molecules. As the Vina score is defined as the negated AutoDock Vina binding free energy, higher scores indicate stronger predicted binding affinity. In each column, \textbf{bold} denotes the top-performing method, while \emph{italics} indicate the second-best. Trio* denotes the model variant without Direct Preference Optimization (DPO), whereas Trio represents the full version trained with DPO.
\end{tablenotes}
\end{threeparttable}
\label{tab:docking_comparison}
\end{table}
\end{landscape}

\subsection*{Target-Specific Molecule Design}
A pivotal task in molecular design is protein-targeted molecule generation, aiming to generate entirely novel compounds exhibiting improved binding affinity toward specific protein targets. Inspired by recent experimental frameworks outlined in \cite{f-rag}, we conducted a comprehensive evaluation of Trio across a series of protein-targeted molecular generation tasks. These tasks involve optimizing the molecular Vina score for five well-established protein targets: PARP1 (Poly [ADP-ribose] polymerase-1), FA7 (Coagulation factor VII), 5HT1B (5-hydroxytryptamine receptor 1B), BRAF (Serine/threonine-protein kinase B-raf), and JAK2 (Tyrosine-protein kinase JAK2). These targets are selected as they encompass diverse protein families and serve as standard benchmarks in recent structure-based drug design literature \cite{geam}. Simultaneously, we assessed whether the generated molecules maintain desirable drug-like properties, quantified using QED, SA, and chemical novelty.

To rigorously benchmark Trio's performance, we compared it against multiple state-of-the-art baseline methods representing various molecular generative strategies. Specifically, the comparative study encompassed four methodological families. Fragment-based approaches, including JT-VAE \cite{jtvae}, HierVAE \cite{hiervae}, MARS \cite{mars}, RationaleRL \cite{rationale}, FREED \cite{freed}, PSVAE \cite{pavae}, f-RAG \cite{f-rag} and GEAM \cite{geam}, construct an explicit fragment vocabulary from molecular data and subsequently assemble these fragments into novel candidates. Genetic-algorithm variants, such as Graph GA \cite{graphga}, GEGL \cite{gegl} and Genetic GFN \cite{gflow}, exploit fragment information through fragment-based crossover operations, whereas GA+D \cite{g+d} applies a discriminator-augmented GA directly to SELFIES representations. RL baselines contain REINVENT \cite{reinvent} which operates on SMILES strings, and MORLD \cite{morld} which utilizes molecular graphs. Finally, MOOD \cite{mood}, a diffusion-based generative model with out-of-distribution control, targets enhanced chemical novelty.

\begin{figure}[htbp]
\centering
\includegraphics[width=\textwidth]{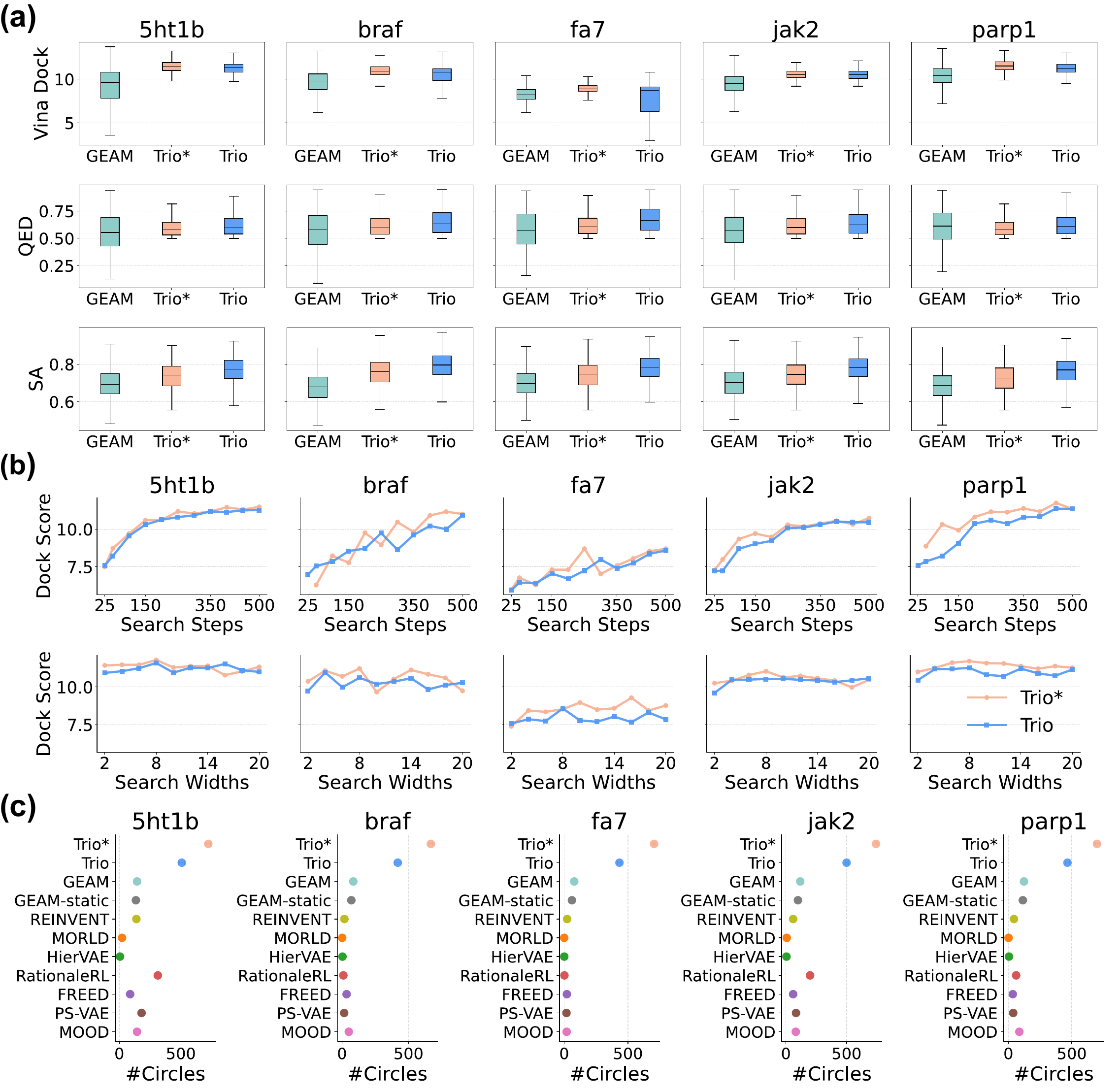}
\caption{\textbf{Performance and Diversity Analysis on Five Therapeutic Targets.} This figure evaluates the effectiveness and diversity of molecules generated by our proposed models, Trio* and Trio, against several baseline methods. \textbf{a}, Box plots comparing the distributions of Vina scores (top), Quantitative Estimate of Drug-likeness (QED, middle), and Synthetic Accessibility (SA, bottom) for molecules generated by GEAM, Trio*, and Trio; \textbf{b}, Hyperparameter sensitivity analysis for Trio* and Trio. The plots show the average Vina score from 20 independent runs as a function of varying search steps (top) and search widths (bottom); \textbf{c}, Molecular diversity analysis using the \#Circles metric. Diversity is quantified by calculating the maximum number of molecules that can be selected from a generated set of 3,000, such that every pair of selected molecules exceeds a minimum distance threshold. A higher \#Circles value signifies greater diversity and exploration of the chemical space.}
\label{fig:mcts}
\end{figure}

\begin{figure}[htbp]
\centering
\includegraphics[width=0.95\textwidth]{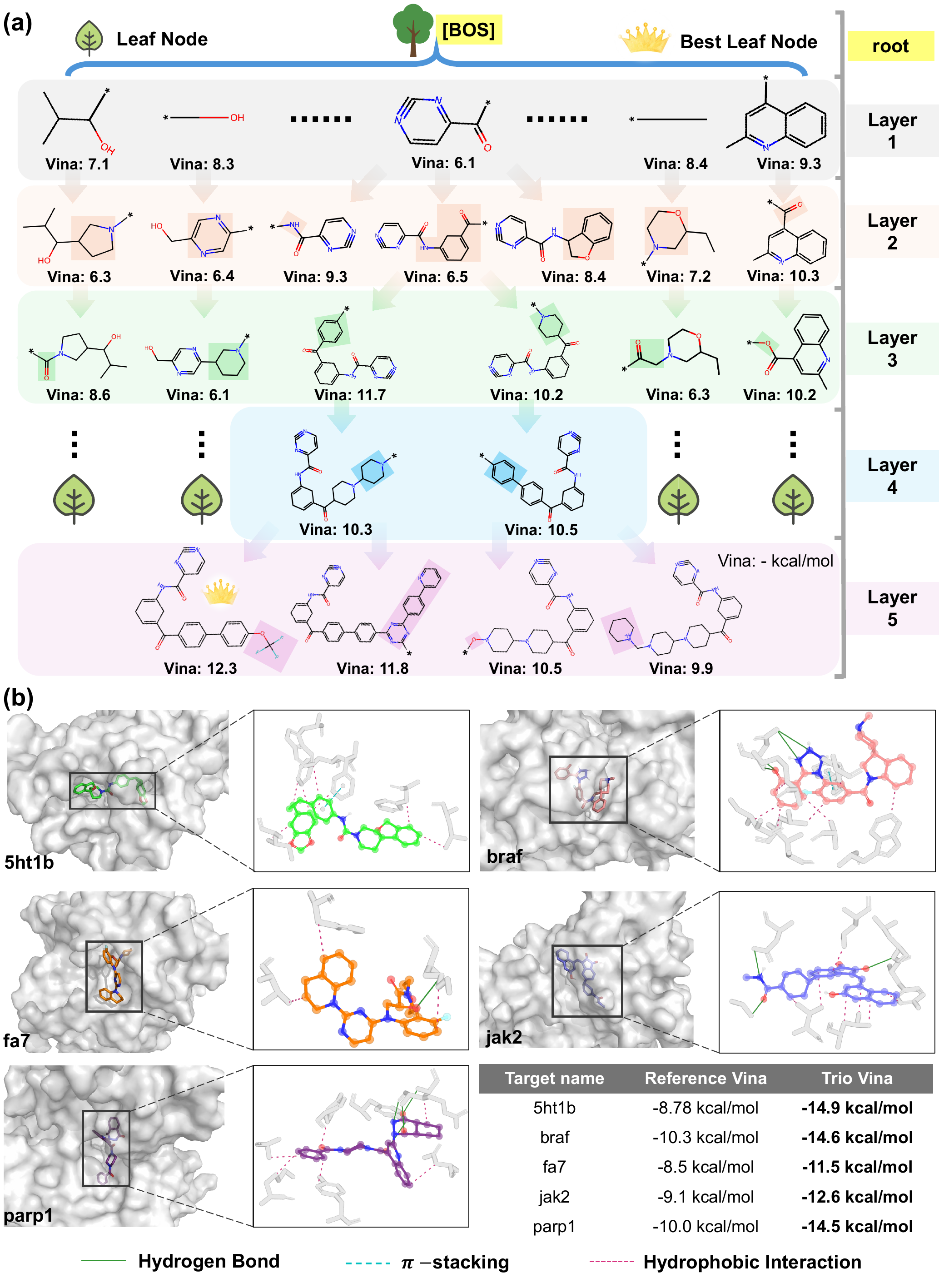}
\caption{\textbf{Illustration of the Trio framework's stepwise generative mechanism and the intermolecular interactions between generated ligands and target protein binding pockets.} \textbf{a}, Schematic illustration of the Monte Carlo Tree Search for target-based \textit{de novo} generation. Starting from the \texttt{[BOS]} root token, molecules are constructed via iterative fragment addition (Layers 1–5) and prioritized by Vina scores to identify the optimal candidate (crown icon); \textbf{b}, Predicted binding modes of generated leads against target proteins. Detailed views of the binding pockets for 5HT1B, BRAF, FA7, JAK2, and PARP1 highlight key non-covalent interactions. Contacts are color-coded: hydrophobic (warmpink dashed), hydrogen bonds (forestgreen solid), and $\pi$–$\pi$ stacking (teal dashed).}
\label{fig:visualization}
\end{figure}

The primary evaluation metric employed in this experiment is the top 5\% hit score, defined as the mean Vina score of the top 5\% of unique and novel generated molecules. To align with a reward-maximization objective, the Vina score is calculated as the negated AutoDock Vina binding free energy; thus, a higher positive value denotes a stronger binding affinity. In contrast to methods such as f-RAG \cite{f-rag} and GEAM \cite{geam}, which are specifically trained or fine-tuned on relatively small-scale datasets such as ZINC250K, FRAGPT is pre-trained on a large-scale molecular corpus without requiring task-specific fine-tuning on curated datasets for downstream adaptation. Accordingly, we identify successful hits based on three stringent criteria to ensure both high binding affinity and favorable drug-likeness: $\text{Vina score} > \text{Vina score}_{\text{active}}^{\text{median}}$, $\text{QED} > 0.5$, and $\text{SA} < 5.0$.

We generated 3,000 candidate molecules for each protein target and benchmarked them against state-of-the-art baseline generators. As shown in Table \ref{tab:docking_comparison}, the foundational Trio* model (operating without DPO constraints) achieves the best binding affinity on five targets, outperforming all baselines. This superiority demonstrates the effectiveness of coupling a generative fragment-level language model with a guided tree search procedure. While previous distribution-learning methods (e.g., \cite{jtvae, hiervae, pavae, mood}) often lack target-oriented generation capability, and rule-constrained algorithms (e.g., GEAM, f-RAG) limit search efficiency, Trio* leverages the MLM's inherent generalization to propose diverse, chemically meaningful fragments. Crucially, MCTS utilizes the MLM's generative probabilities to intelligently prioritize exploration paths most likely to improve docking scores, enabling Trio* to converge efficiently toward high-affinity candidates without relying on rigid heuristics.

Building on the robust search capability of Trio*, the full Trio framework integrates FRAGPT-DPO with MCTS to create a holistic solution for drug-like molecule search. Unlike the exploration-focused Trio*, the full Trio model does not optimize solely for binding affinity but navigates a multi-objective landscape to prioritize drug-likeness and synthetic accessibility. Consequently, Trio maintains a competitive Vina score relative to previous state-of-the-art methods while significantly enhancing pharmacological properties. To mitigate inflated performance caused by clusters of near-identical molecules, we compared 3,000 molecules per target generated by Trio*, Trio, and GEAM (chosen as the strongest open-source baseline, since newer models like f-rag are not publicly available). We then computed the Morgan-Tanimoto similarity coefficient to discard any pair with a similarity greater than 0.4. Because GEAM optimizes within the limited ZINC250K database, nearly half of its molecules are removed. In contrast, even after removing structurally redundant pairs, both Trio* and Trio retained over 70\% of their generated candidates, highlighting their generative breadth. As summarized in Fig.~\ref{fig:mcts}a, while Trio* exhibits the most extreme Vina score distribution, the full Trio model achieves superior and tightly clustered values for QED and SA driven by preference alignment, offering the optimal balance for practical drug discovery.

To assess how the number of Monte-Carlo simulations and the tree width affect MCTS performance, we analyzed the resulting docking scores across varying simulation counts and tree widths. As shown in Fig.~\ref{fig:mcts}b, docking scores generally improve as the simulation count increases. By contrast, expanding the tree width enhances exploration but yields no statistically significant gain in docking performance.

To further assess chemical space coverage of generated molecules, we adopt the \#Circles metric as introduced in \cite{xie2023much}, which identifies distinct chemical clusters by iteratively removing molecules with high similarity ($\text{Morgan-Tanimoto similarity} > 0.75$). As shown in Fig.~\ref{fig:mcts}c, conventional distribution-learning models (e.g., HierVAE and MORLD) tend to suffer from mode collapse, generating outputs clustered near the training distribution. Although hybrid approaches such as GEAM incorporate a genetic algorithm to refine candidate molecules, they remain fundamentally constrained by predefined fragment libraries. This reliance on static building blocks inherently limits the diversity of generated molecules, restricting exploration to the chemical space spanned by the initial library. In contrast, the Trio* model demonstrates a significant multi-fold improvement in \#Circles across all five protein targets, reflecting its unconstrained capacity for exploration. The full Trio model exhibits an expected moderate reduction relative to Trio due to the constraints of preference alignment, yet its \#Circles count remains superior to earlier methods. This significant boost highlights the ability of our method to overcome the limitations of rule-based search and static fragment libraries, enabling more diverse and novel molecular generation. Crucially, our advantage is consistent across all targets regardless of receptor type or structural complexity, suggesting that the combination of MLM and tree search robustly generalizes across different biological contexts. This consistency circumvents the target transferability issues frequently observed in purely data-driven or rule-constrained methods, demonstrating the unique adaptability of Trio in navigating diverse chemical landscapes.

Trio’s synergistic architecture is depicted as an MCTS-guided hierarchical search procedure in Fig.~\ref{fig:visualization}a, which seamlessly integrates the extensive semantic knowledge encoded in FRAGPT with the established search efficiency of MCTS. Specifically, pretrained on large-scale and unlabeled molecular corpora, FRAGPT can serve as a dynamic and chemically expressive source of fragment proposals, in sharp contrast to methods limited by static, predefined fragment libraries. By capturing the syntactic and semantic regularities of chemical structures, it generates fragments that are simultaneously diverse, synthetically plausible, and chemically coherent. MCTS then navigates this expansive LM-generated chemical space, striking an effective balance between exploitation (intensively refining trajectories that yield high docking scores) and exploration (probing less obvious regions to avoid local optima and uncover novel scaffolds). This navigation is jointly informed by the MLM’s generative likelihoods and real-time docking evaluations, thereby establishing a robust, closed-loop framework that tightly couples generative language modeling with target-aware molecular optimization.

The comprehensive visualization of the entire search tree affords a level of interpretability rarely attained in \textit{de novo} molecular design. By explicitly charting the generative trajectory, Fig.~\ref{fig:visualization}a enables researchers to systematically trace the evolutionary lineage of candidate molecules, revealing the stepwise incorporation of chemical features and structural motifs that recurrently enhance predicted binding affinity. This granular transparency transcends the mere presentation of final optimized compounds, providing a mechanistic framework that elucidates how specific functional groups and fragment combinations contribute to ligand potency. Moreover, it highlights fragment-level binding propensities within diverse chemical contexts, thereby deepening our understanding of structure–activity relationships. Collectively, this interpretable design paradigm empowers medicinal chemists with actionable insights, offering a more rational, human-in-the-loop workflow that bridges generative modeling with expert-driven drug discovery.

Complementing the trajectory analysis, Fig.~\ref{fig:visualization}b substantiates Trio’s target-aware \textit{de novo} design capability through structural validation across diverse protein binding pockets. The interaction analyses reveal that these compounds achieve exceptionally favorable predicted binding free energies and engage in key noncovalent interactions, such as directional hydrogen bonds, $\pi$–$\pi$ stacking, and hydrophobic contacts. Notably, a comparative table within Fig.~\ref{fig:visualization}b demonstrates that the predicted binding affinities of Trio-generated ligands substantially surpass those of reference compounds across multiple target pockets, with an average increase of 46.0\%. This comprehensive interaction profiling corroborates the model’s ability to generate chemically valid, synthetically accessible ligands with enhanced specificity and predicted affinity.

Collectively, the robustness of Trio can be attributed to the fact that a fragment-level language model provides chemically meaningful and highly diverse candidates, and a target-aware search loop (e.g., MCTS) guided by docking feedback selectively amplifies pathways that improve binding affinity in real time. This closed loop simultaneously broadens the accessible chemical space and accelerates convergence to high-quality candidates, compared to earlier pipelines that relied on static fragment libraries, hand-crafted genetic operators, or one-shot scoring. As such, the consistently superior docking energies highlight not only the effectiveness of FRAGPT itself, but also the broader promise of combining large-scale chemical language models with adaptive search as a general approach for target-specific molecular design.

\begin{figure}[htbp]
\centering
\includegraphics[width=\textwidth]{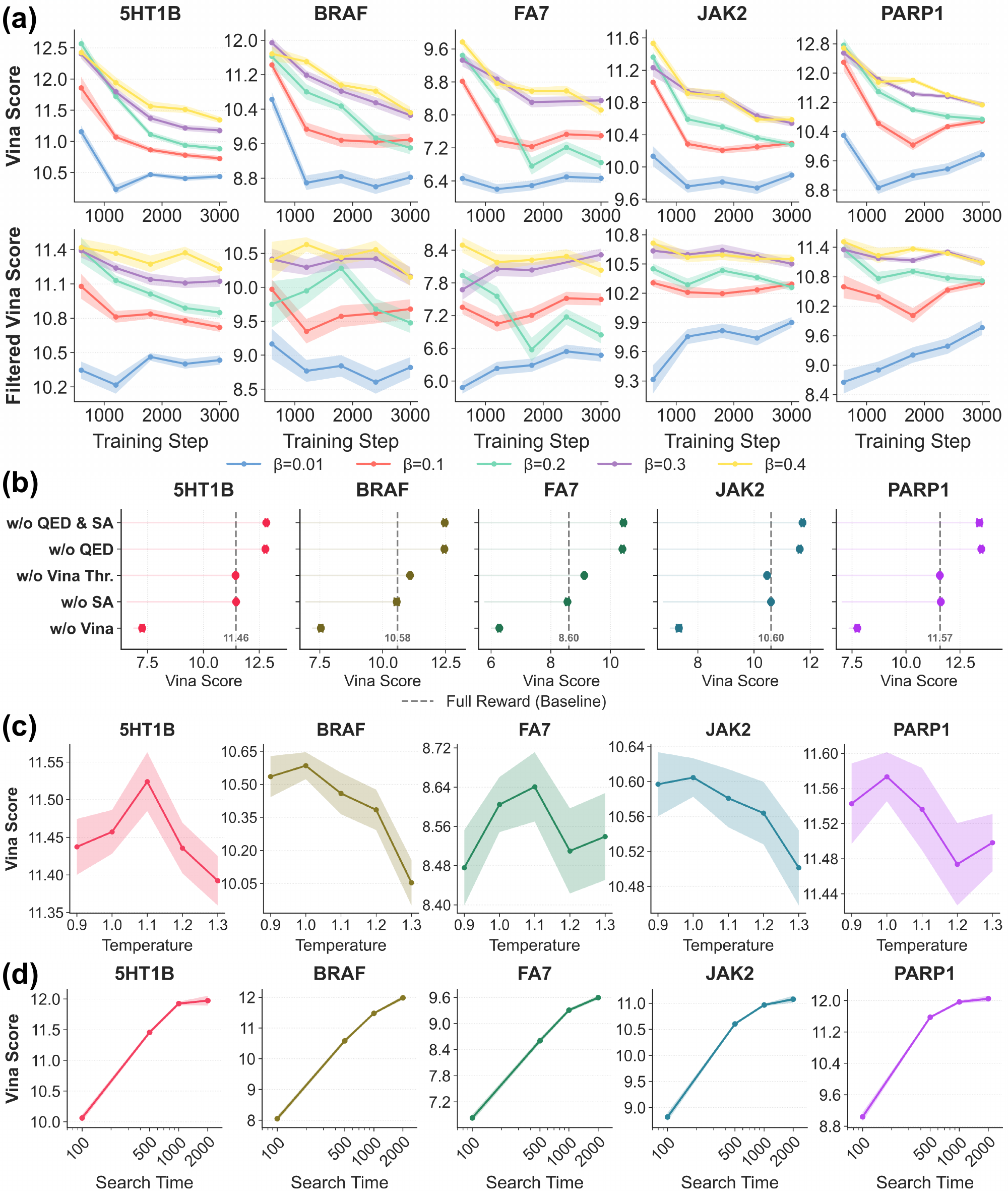}
\caption{\textbf{Evaluation of optimization dynamics, reward formulation, and scaling properties in the MCTS-guided molecular generation framework.} \textbf{a}, Influence of the Direct Preference Optimization (DPO) hyperparameter $\beta$ on model training trajectories across five protein targets. Raw (top) and filtered (bottom) Vina scores (negated binding free energies) are tracked across 3,000 steps for $\beta$ values ranging from 0.01 to 0.4, illustrating the trade-off between binding affinity exploration and physicochemical constraint adherence; \textbf{b}, Ablation study of the multi-objective reward function. The panel displays the impact of individually removing constraints (e.g., Vina score, QED, SA) on the optimization boundaries and final binding affinity; \textbf{c},Sensitivity of generative performance to the language model's sampling temperature, highlighting the optimal temperature range (1.0–1.1) for maximizing docking scores before stochasticity degrades performance; \textbf{d}, Scaling behavior of the algorithm based on MCTS search time. The plot shows the monotonic improvement and subsequent convergence of Vina scores as search steps increase from 100 to 2,000.}
\label{fig:mcts_extra}
\end{figure}

\subsection*{Monte Carlo tree search and preference optimization guide structurally valid molecular design}

To evaluate the optimization dynamics, reward formulation, and inference-time scaling properties of our Monte Carlo Tree Search (MCTS) guided generation framework, we conducted experiments across five protein targets: 5HT1B, BRAF, FA7, JAK2, and PARP1. 

We first investigated the influence of the DPO hyperparameter $\beta$ by evaluating intermediate checkpoints at 600-step intervals up to 3,000 training steps (Fig.~\ref{fig:mcts_extra}a). In this context, DPO is primarily utilized to align the generation process with favorable drug-like properties, specifically the quantitative estimate of drug-likeness (QED) and the synthetic accessibility (SA) score. For each target, we generated 150 molecules for each $\beta \in [0.01, 0.4]$ and tracked both the raw and filtered AutoDock Vina scores. The filtered Vina score is averaged exclusively over molecules that satisfy strict physicochemical constraints, defined as having a normalized QED and a normalized SA score both strictly greater than 0.5.

The parameter $\beta$ scales the Kullback-Leibler (KL) divergence penalty, effectively controlling the deviation from the reference policy. A larger value of $\beta$ constrains the generation policy to remain closer to the reference distribution. Because this reference distribution is not strictly bounded by predefined QED and SA thresholds, it retains a broader exploration capacity within the chemical space to discover high binding affinities. This algorithmic property accounts for the elevated raw Vina scores observed at larger $\beta$ values (e.g., $\beta=0.4$), a trend consistent with the extensive exploration capabilities demonstrated in Table~\ref{tab:docking_comparison}. Notably, for $\beta = 0.01$, the raw Vina scores remain at inherently suboptimal levels, despite exhibiting a slight upward trend during the training process. This phenomenon occurs because a substantially low penalty ($\beta = 0.01$) permits excessive deviation from the reference distribution, thereby compromising both the validity of the generated molecules and the overall exploration efficacy.

Furthermore, analyzing these metrics across the training steps revealed a critical dynamic regarding the trade-off between binding affinity and drug-likeness. During the early stages of training, the models heavily leverage the unconstrained search space of the reference distribution, initially yielding highly favorable raw Vina scores. However, as the training progresses toward the 3,000-step mark, a significant divergence in model performance emerges across different $\beta$ settings. As DPO increasingly enforces strict physicochemical constraints, the overall raw Vina scores experience a consistent reduction across all configurations. For larger $\beta$ values, the strong KL penalty restricts deviation from the reference model, which mitigates this reduction and enables the raw Vina scores to remain significantly elevated. Concurrently, the filtered Vina scores under these larger $\beta$ settings tend to plateau, reflecting the inherent chemical difficulty of simultaneously maintaining strong binding affinities while satisfying strict QED and SA boundaries (both $> 0.5$). Ultimately, the filtered Vina score trajectories indicate that an appropriately calibrated $\beta$ is required to effectively balance the broad exploration capacity with the necessary physicochemical constraints throughout the training process.

To evaluate our multi-objective design, we performed an ablation study on the reward function components (Fig.~\ref{fig:mcts_extra}b). Here, we generated 300 molecules per configuration for each target using the base model without DPO fine-tuning. Removing individual constraints altered the model's optimization behavior. Omitting the primary docking reward (w/o Vina) deprived the search process of its directional signal, resulting in significantly less favorable Vina scores. Conversely, removing the physicochemical constraints (w/o QED \& SA) relaxed the optimization boundaries, thereby increasing the algorithm's search capability for raw binding affinity. Furthermore, ablating the Vina score threshold (w/o Vina threshold) demonstrated that this constraint primarily modulates the stringency of the search process toward the target score. Excluding only the SA constraint (w/o SA) yielded a negligible impact on the optimization trajectory. This minimal deviation indicates that the base model intrinsically generates molecules with high synthetic accessibility, reflecting an inherent structural validity that corroborates the findings in Fig.~\ref{fig:rl}b.

We next assessed the sensitivity of the generative process to the language model's sampling temperature (Fig.~\ref{fig:mcts_extra}c), generating 300 molecules per target for each setting. The Vina scores exhibited a non-linear trend. Temperatures between 1.0 and 1.1 provided the optimal balance between exploration and exploitation, yielding the highest docking scores. Conversely, higher temperatures (e.g., 1.2 and 1.3) introduced excessive stochasticity, disrupting learned chemical semantics and leading to a sharp decline in performance.

Finally, we performed an extended ablation on the MCTS search time to evaluate the algorithm's scaling behavior (Fig.~\ref{fig:mcts_extra}d). By increasing the search steps from 100 to 2,000 (generating 300 molecules per setting), we observed that the Vina scores improved monotonically across all targets. The most significant performance gains are achieved during the initial expansion from 100 to 1,000 steps. Beyond this point, improvements began to plateau as the tree search converged on the highest-affinity candidates within the constrained chemical subspace, demonstrating the scalability of the MCTS integration.

% *** Conclusion ***
\section*{Discussion}
Our results demonstrate that combining fragment-level language modeling with tree-based search provides a viable framework for target-aware de novo molecular design across the evaluated benchmarks. In Trio, targeted generation is framed not only as distribution modeling but also as a sequential decision process wherein chemically plausible fragment choices are guided by target-specific objectives during the search.

Within this framework, FRAGPT serves as a local proposal model for fragment assembly, whereas MCTS reweights these proposals according to a target-specific reward. This coupling of local chemical plausibility with global search distinguishes Trio from pipelines in which generation and optimization are performed separately. At the same time, the current implementation remains limited by the fact that the language model is target-agnostic at inference, with target information entering only through the reward function. It remains to be determined whether lightweight pocket-aware conditioning could bias fragment prediction toward more compatible chemotypes and thereby reduce the search budget required to identify strong candidates.

A principal limitation of this approach lies in the fidelity of the reward function. In the current implementation, the MCTS is guided by AutoDock Vina docking scores. Although Vina is efficient and widely used, it does not fully account for protein flexibility, induced fit, solvation, and entropic effects. Consequently, the search process may favor molecules that exploit scoring-function artifacts rather than those possessing genuinely favorable binding affinities. This limitation is ubiquitous in structure-based molecular generation and is frequently amplified by iterative search algorithms. More informative reward functions, including learned affinity predictors or rigorous free-energy approximations, could mitigate this issue, albeit at the expense of increased computational cost.

The search tree provides trajectory-level interpretability by exposing fragment assembly paths associated with high predicted binding scores. This may help identify which fragment choices are repeatedly selected within a given reward landscape. However, the current search tree is generative rather than retrosynthetic, meaning that the identified assembly paths are not guaranteed to correspond to synthetically accessible routes. Incorporating synthetic feasibility as an auxiliary objective would make the search output more relevant to downstream medicinal chemistry applications.

A further practical concern is computational scalability, as search cost grows rapidly with problem complexity. Specifically, this expense scales with tree depth, the number of expanded states, and the per-node cost of reward evaluation. This computational burden will become more prohibitive if the framework is extended to multifaceted objectives, such as selectivity, uncertainty quantification, or synthesis constraints. Although learned value models, adaptive search budgets, and parallel reward evaluations may partially mitigate this challenge, the efficiency of Trio under more complex reward settings remains to be established.

Overall, the present study indicates that fragment-level language modeling and tree search can function as complementary components rather than competing alternatives. Their integration offers a promising strategy to combine broad chemical exploration with target-directed optimization, although future progress will depend on advanced reward design, enhanced target conditioning, and highly efficient search algorithms.

% *** METHODS ***
\section*{Methods}
\label{methods}

\subsection*{Datasets}
The ZINC \cite{zinc} and UniChem \cite{unichem} databases collectively contain over 1.1 billion SMILES strings. From these, SAFE \cite{safegpt} carefully constructs a diverse set of molecule types into a unified database, spanning drug-like compounds, peptides, multi-fragment molecules, polymers, reagents, and non-small molecules. To reduce the computational burden, only about 15 million SMILES strings are randomly sampled for training Trio from the SAFE dataset without applying additional data augmentation techniques.

Following the methodology of Marco et al. \cite{podda2020deep}, we applied the Breaking of Retrosynthetically Interesting Chemical Substructures (BRICS) algorithm to molecules, fragmenting each from left to right into multiple FragSeqs (Fig.~\ref{fig:fragpt}a). BRICS defines 16 chemical environments that flexibly determine suitable bond cleavage sites and retained functional groups (e.g., aromatic rings and cyclic structures), thereby generating a variety of synthetically feasible fragments \cite{degen2008art}. To facilitate the reconstruction of complete compounds from these fragments, BRICS attaches dummy atoms at each cleavage site, marking the points at which fragments can be rejoined. By following these cleavage labels, the original molecular structure can be reconstructed from the FragSeq. In total, this dataset contains around 10 million FragSeqs.

\subsection*{FRAGPT Architecture}
FragSeqs consist of sequentially arranged SMILES-based fragments, from which the complete molecular SMILES strings can be reconstructed. Structurally, these FragSeqs closely resemble natural language sentences, enabling the application of advanced NLP techniques for chemical structure generation. To tokenize the FragSeqs, the model employs a widely-used tokenizer based on a regular expression pattern tailored for SMILES syntax \cite{schwaller2019molecular}. This tokenizer is designed to capture the atomic and functional group semantics inherent in SMILES strings. The resulting vocabulary consists of approximately 600 unique tokens, encompassing not only standard chemical tokens (e.g., atoms, bonds, branches, and ring symbols) but also all required special tokens such as \texttt{[BOS]} (Beginning-Of-Sequence), \texttt{[EOS]} (End-Of-Sequence), \texttt{[SEP]} (Fragment identifiers) and \texttt{[PAD]} (Padding indicators). Such a tokenization scheme ensures consistent encoding of molecular fragments while preserving syntactic validity.

To exploit the state-of-the-art performance of current language models, we developed a GPT-based MLM, termed FRAGPT, which is a decoder-only transformer architecture comprising 87.3 million parameters. Its architecture is specifically tailored for the generation of FragSeqs. First, the input sequence tokens are linearly projected into Query ($Q$), Key ($K$), and Value ($V$) matrices. Subsequently, the standard self-attention mechanism computes the dot product between the Query and Key matrices, and the resulting scores are normalized via a softmax function to produce attention weights for the Value matrix. Formally, given input embeddings $\mathbf{X}\in \mathbb{R}^{n \times d}$, the computational procedure can be formally represented as follows:
\begin{align}
\mathrm{Attention}(Q,K,V)= \mathrm{softmax}\left(\frac{QK^T}{\sqrt{d_k}}\right)V,
\end{align}
where $T$ indicates transposition. $Q=XW_Q$, $K=XW_K$ and $V=XW_V$ represent learned projections of the input embeddings $X$. A feed-forward network is then applied to the resulting output embeddings to obtain the semantic features of each token. This attention mechanism enables contextual modeling of fragment interactions through dot product operations.

During the training phase, FragSeqs are first embedded into feature matrices, after which positional embeddings (implemented via Rotary Position Embedding, RoPE) are incorporated \cite{su2024roformer}. The combined embeddings, denoted as $X$, are then fed into the FRAGPT model. The output of FRAGPT is processed through a softmax layer, where each token predicts the next token via a masking mechanism designed to prevent information leakage from future positions. The training objective is to minimize the cross-entropy loss between the model's predicted token probability distribution and the true distribution of target tokens. Formally, given a FragSeq $Y=\{y_1,y_2,\dots,y_t\}$, the loss function can be described as:
\begin{align}
L = -\sum_{t=1}^{T} \log P(y_t|x, y_{<t}),
\end{align}
where $P(y_t|x, y_{<t})$ denotes the conditional probability of the target token $y_t$, given the input embedding $x_t$ and all preceding tokens $y_{<t}$. FRAGPT is trained using the AdamW optimizer with hyperparameters $\beta_1=0.9$ and $\beta_2=0.95$. The training is conducted on six NVIDIA A6000 GPUs for a total of 8 epochs, employing a learning rate scheduling strategy that combines an initial warm-up phase with a subsequent linear decay. The batch size is set to 32 samples per GPU, ensuring stable optimization and efficient utilization of computational resources.

\subsection*{Reinforcement Learning}
Inspired by the success of GPT models enhanced through reinforcement learning (RL) algorithms for a variety of NLP tasks, we employ FRAGPT as the base model and integrate different RL approaches tailored to each downstream task. This strategy leverages the strengths of RL-based optimization to further improve the performance and adaptability of the base model in specific applications.

\textbf{Direct Preference Optimization}: To encourage FRAGPT to generate more reasonable molecules, we adopt the DPO algorithm to smoothly align the model towards higher QED and lower SA, instead of using Augmented Likelihood Reinforcement Learning, which collapses the output distribution into peaky modes over desirable properties \cite{olivecrona2017molecular}. Unlike Proximal Policy Optimization (PPO) \cite{ppo}, which requires training an auxiliary reward model, DPO treats the GPT policy as the reward model. This design yields an explicit mapping between policy logits and reward signals, allowing the language model to satisfy user-defined preferences without extra critics. The general DPO pipeline employed in our experiments can be summarized as follows.

\begin{itemize}
\item[$\bullet$] Generate FragSeqs using the policy initialized with FRAGPT $(\pi_\text{ref})$:
\begin{align}
y \sim \pi_\text{ref}(\cdot \mid x),
\end{align}
where $y$ denotes a generated FragSeq, and $x$ represents its prior fragment context, i.e., the sequence of fragments generated up to that point. Specifically, we group each FragSeq by its prefix fragment and restrict every molecule to contain no more than three identical prefix fragments. When a group lacks a common prefix fragment or contains fewer than 8 molecules, we assign \texttt{[BOS]} as the default prefix fragment. Next, we annotate each FragSeq with its calculated pharmacological attributes (e.g., QED and SA). To construct meaningful preference pairs, we first rank the molecules in each group by their drug attributes. We then draw positive-negative pairs from the top and bottom of each ranking, thus obtaining an offline preference dataset that shares a common fragment prefix yet spans a broad range of difficulty levels:
\begin{align}
\mathcal{D} = \{(x^{(i)}, y_{g}^{(i)}, y_{l}^{(i)})\}_{i=1}^N,
\end{align}
where $y_g^{(i)}$ and $y_l^{(i)}$ denote the FragSeqs derived from the same prior fragment $x^{(i)}$ that exhibit higher and lower drug-property scores, respectively. This DPO dataset construction method is specifically tailored to FragSeqs and closely mirrors the original DPO procedure used in NLP. The initial fragment sequences act as prompts, enabling the model to distinguish the relative drug-property quality of two molecules derived from the same starting fragment.

\item[$\bullet$] Maximize the likelihood of the reinforced MLM $\pi_\theta$ with respect to the reference policy $\pi_\text{ref}$, where the optimization objective is given by:
\begin{align}
\mathcal{L}_\text{DPO} = - \mathbb{E}_{(x,y_g,y_l) \sim \mathcal{D}}\bigg[ \log \sigma \bigg( \beta \log \frac{\pi_\theta(y_g|x)}{\pi_\text{ref}(y_g|x)} - \beta \log \frac{\pi_\theta(y_l|x)}{\pi_\text{ref}(y_l|x)} \bigg)\bigg],
\end{align}
where $\sigma$ is the sigmoid function, and $\beta$ is a scaling coefficient that adjusts the trade-off between enhancing preference and preserving the original distribution during training.
\end{itemize}

\subsection*{Monte Carlo Tree Search}
Monte Carlo Tree Search (MCTS) is a heuristic search algorithm renowned for its efficacy in sequential decision-making tasks, particularly in domains requiring combinatorial optimization, most notably in AlphaGo \cite{chaslot2008monte}. In the context of molecule generation, MCTS operates as an iterative, tree-based framework that balances exploration of potential chemical spaces with exploitation of promising molecular candidates. The algorithm comprises four canonical phases: Selection, Expansion, Simulation and Backpropagation (Fig.~\ref{fig:overall}b). It iteratively grows a decision tree to guide fragment assembly by searching for the optimal fragment sequence, identifying the most promising molecular candidates. Below, we elaborate on each phase and its corresponding role in the molecular generation pipeline:
\begin{itemize}
\item[$\bullet$]\textbf{Selection Phase}: Navigating the Chemical Decision Tree. The algorithm begins with a predefined root node, which may represent a \texttt{[BOS]} token or an initial molecular fragment encoded in SMILES notation. The selection strategy employs a modified Upper Confidence Bound for Trees (UCT) criterion to select a child node with high potential rewards while maintaining diversity in exploration \cite{auer2002finite}. The UCT value for a child node $j$ is formulated as:
\begin{align}
UCT_{j} = \alpha \times \mathrm{average}(a_j) + (1-\alpha)\times\max(a_j) + C \sqrt{\frac{\ln N_C}{N_j}},
\end{align}
where $\mathrm{average}(a_j)$ and $\max(a_j)$ represent the average and maximum rewards of action $a_j$, respectively. $\alpha$ manipulates the trade-off between historical performance $\mathrm{average}(a_j)$ and optimistic potential $\max(a_j)$, and $C$ represents the exploration-exploitation balance by scaling the second term derived from the UCT framework. $N_C$ is the total visitation count of the parent node, and $N_j$ is the visitation count of node $j$. This dual-objective formulation ensures that under-explored nodes with high variance in rewards receive attention, thereby mitigating premature convergence to suboptimal regions.

To further investigate the potential nodes, a children-adaptive strategy, as proposed by Tian et al. \cite{tian2024toward}, is employed to dynamically adjust the branching factor of nodes based on their reward stability. The importance metric $I(s_t)$ for the node $s_t$ is calculated as:
\begin{align}
I(s_t) = \max_{o_t^i} |R(s_t, o_t^i) - \bar{R(s_t)}|,
\end{align}
where $R(s_t, o_t^i)$ is the reward of the $i$-th child of $s_t$, and $\bar{R(s_t)}$ is the mean reward across all children. Intuitively, a high $I(s_t)$ indicates significant reward deviation among children, promoting the algorithm to expand the number of child nodes to $n(s_t) = \min\big(\beta \lfloor I(s_t) \rfloor, c_{\max}\big)$, where $\beta$ scales the expansion rate and $c_{\max}$ imposes an upper bound to prevent computational overload. This adaptive mechanism ensures that nodes with fluctuating reward distributions require deeper exploration, enhancing the likelihood of discovering high-reward molecular candidates.

\item[$\bullet$]\textbf{Expansion Phase}: Probabilistic Generation of Next Molecular Fragments. After selecting a leaf node, the algorithm first evaluates its terminal status. If the leaf node contains an \texttt{[EOS]} token, the process returns to the selection phase. Otherwise, FRAGPT acts as the agent to generate the subsequent fragment of the SMILES sequence, conditioned on the current molecular state derived from the chemical context of the parent node. During this phase, FRAGPT generates only the next fragment SMILES by expanding a single branch from the selected node, rather than producing the entire sequence until the \texttt{[EOS]} token is reached. The generated fragment is appended to the current SMILES string, extending the molecular context and creating a new child node in the decision tree. More importantly, the expansion stage incorporates a duplicate detection mechanism, which calculates the molecular similarity between the current node and its sibling nodes. To avoid redundant exploration, the expansion is repeated up to five times until a structurally distinct molecule is obtained, thereby enhancing both the diversity of candidates and the overall efficiency of the optimization process.

\item[$\bullet$]\textbf{Simulation Phase}: Rollout Strategies and Reward Estimation. The simulation phase evaluates the long-term potential of the newly expanded node by performing Monte Carlo rollouts until a terminal state (\texttt{[EOS]}) is reached. During the rollout process, FRAGPT generates the integral candidate SMILES strings based on the current node state. In contrast to the expansion phase, the simulation phase treats FRAGPT as a simulator that generates the complete SMILES sequence and reconstructs the corresponding molecule, approximating the potential molecular state of the currently expanded node for subsequent evaluation. The resulting molecule is scored using a domain-specific reward function $R(\cdot)$, which quantifies desirable properties such as synthetic accessibility (SA), quantitative estimate of drug-likeness (QED), and target-specific bioactivity (e.g., docking scores). To enhance robustness, the reward function can be designed to incorporate ensemble evaluations, integrating multiple scoring functions that reflect diverse molecular objectives, facilitating more reliable assessments and supporting multi-objective optimization during molecular generation.

\item[$\bullet$]\textbf{Backpropagation Phase}: Reward Dissemination and Tree Updates. The final reward $R$ obtained from the simulation is propagated backward through the tree to update the statistics of all traversed nodes. Each node's visitation count $N_j$ and cumulative reward $Q_j$ are incremented as:
\begin{align}
N_j \leftarrow N_j + 1, \quad Q_j \leftarrow Q_j + R.
\end{align}
This update mechanism enables the algorithm to accumulate experience over time, reinforcing nodes that consistently lead to high-reward outcomes while gradually discouraging exploration of suboptimal branches. By aggregating reward information in this manner, the tree progressively biases future selections toward more promising regions of the molecular space, improving search efficiency and optimization performance.
\end{itemize}

The MCTS algorithm initializes with a root node defined by task-specific constraints (e.g., a scaffold structure or \texttt{[BOS]} token) and iteratively cycles through the four phases until a termination condition is met. In this study, the typical termination criterion is a predefined number of MCTS iterations (Iteration Limit). After optimization, the optimal molecule is selected from the leaf nodes based on the highest cumulative reward, with optional post-processing (e.g., validity checks) to refine the output.

\subsection*{Code availability}
The source code of Trio is available at Github: \url{https://github.com/SZU-ADDG/Trio}.

\subsection*{Data availability}
The data that support the findings of this study are available from the following sources: The training dataset was derived by sampling molecules from the ZINC database, which is publicly accessible at \url{https://zinc.docking.org/}. Datasets used for evaluation are obtained from public databases. All data related to the specific targets, along with the source code, are provided in the aforementioned GitHub repository.

% *** REFERENCES ***
% \newpage
\bibliography{bibliography}

@article{kopf2023openassistant,
  title={Openassistant conversations-democratizing large language model alignment},
  author={K{\"o}pf, Andreas and Kilcher, Yannic and Von R{\"u}tte, Dimitri and Anagnostidis, Sotiris and Tam, Zhi Rui and Stevens, Keith and Barhoum, Abdullah and Nguyen, Duc and Stanley, Oliver and Nagyfi, Rich{\'a}rd and others},
  journal={Advances in Neural Information Processing Systems},
  volume={36},
  pages={47669--47681},
  year={2023}
}

@article{sun2025synllama,
  title={SynLlama: Generating Synthesizable Molecules and Their Analogs with Large Language Models},
  author={Sun, Kunyang and Bagni, Dorian and Cavanagh, Joseph M and Wang, Yingze and Sawyer, Jacob M and Zhou, Bo and Gritsevskiy, Andrew and Zhang, Oufan and Head-Gordon, Teresa},
  journal={ACS Central Science},
  year={2025},
  publisher={ACS Publications}
}

@article{krishnan2025generative,
  title={A generative deep learning approach to de novo antibiotic design},
  author={Krishnan, Aarti and Valeri, Jacqueline A and Jin, Wengong and Donghia, Nina M and Sieben, Leif and Luttens, Andreas and Zhang, Yu and Modaresi, Seyed Majed and Hennes, Andrew and Fromer, Jenna and others},
  journal={Cell},
  year={2025},
  publisher={Elsevier}
}

@article{zhang2025artificial,
  title={Artificial intelligence in drug development},
  author={Zhang, Kang and Yang, Xin and Wang, Yifei and Yu, Yunfang and Huang, Niu and Li, Gen and Li, Xiaokun and Wu, Joseph C and Yang, Shengyong},
  journal={Nature medicine},
  volume={31},
  number={1},
  pages={45--59},
  year={2025},
  publisher={Nature Publishing Group US New York}
}

@article{zitnik2025ai,
  title={AI-enabled drug discovery reaches clinical milestone: Machine learning},
  author={Zitnik, Marinka},
  journal={Nature Medicine},
  pages={1--2},
  year={2025},
  publisher={Nature Publishing Group US New York}
}

@article{wu2024tamgen,
  title={TamGen: drug design with target-aware molecule generation through a chemical language model},
  author={Wu, Kehan and Xia, Yingce and Deng, Pan and Liu, Renhe and Zhang, Yuan and Guo, Han and Cui, Yumeng and Pei, Qizhi and Wu, Lijun and Xie, Shufang and others},
  journal={Nature Communications},
  volume={15},
  number={1},
  pages={9360},
  year={2024},
  publisher={Nature Publishing Group UK London}
}

@article{noutahi2024gotta,
  title={Gotta be SAFE: a new framework for molecular design},
  author={Noutahi, Emmanuel and Gabellini, Cristian and Craig, Michael and Lim, Jonathan SC and Tossou, Prudencio},
  journal={Digital Discovery},
  volume={3},
  number={4},
  pages={796--804},
  year={2024},
  publisher={Royal Society of Chemistry}
}

@article{krenn2020self,
  title={Self-referencing embedded strings (SELFIES): A 100\% robust molecular string representation},
  author={Krenn, Mario and H{\"a}se, Florian and Nigam, AkshatKumar and Friederich, Pascal and Aspuru-Guzik, Alan},
  journal={Machine Learning: Science and Technology},
  volume={1},
  number={4},
  pages={045024},
  year={2020},
  publisher={IOP Publishing}
}

@article{weininger1988smiles,
  title={SMILES, a chemical language and information system. 1. Introduction to methodology and encoding rules},
  author={Weininger, David},
  journal={Journal of chemical information and computer sciences},
  volume={28},
  number={1},
  pages={31--36},
  year={1988},
  publisher={ACS Publications}
}

@article{wang20253dsmiles,
  title={3DSMILES-GPT: 3D molecular pocket-based generation with token-only large language model},
  author={Wang, Jike and Luo, Hao and Qin, Rui and Wang, Mingyang and Wan, Xiaozhe and Fang, Meijing and Zhang, Odin and Gou, Qiaolin and Su, Qun and Shen, Chao and others},
  journal={Chemical Science},
  volume={16},
  number={2},
  pages={637--648},
  year={2025},
  publisher={Royal Society of Chemistry}
}

@inproceedings{zholus2025bindgpt,
  title={Bindgpt: A scalable framework for 3d molecular design via language modeling and reinforcement learning},
  author={Zholus, Artem and Kuznetsov, Maksim and Schutski, Roman and Shayakhmetov, Rim and Polykovskiy, Daniil and Chandar, Sarath and Zhavoronkov, Alex},
  booktitle={Proceedings of the AAAI Conference on Artificial Intelligence},
  volume={39},
  number={24},
  pages={26083--26091},
  year={2025}
}

@article{du2024machine,
  title={Machine learning-aided generative molecular design},
  author={Du, Yuanqi and Jamasb, Arian R and Guo, Jeff and Fu, Tianfan and Harris, Charles and Wang, Yingheng and Duan, Chenru and Li{\`o}, Pietro and Schwaller, Philippe and Blundell, Tom L},
  journal={Nature Machine Intelligence},
  volume={6},
  number={6},
  pages={589--604},
  year={2024},
  publisher={Nature Publishing Group UK London}
}

@article{song2023equivariant,
  title={Equivariant flow matching with hybrid probability transport for 3d molecule generation},
  author={Song, Yuxuan and Gong, Jingjing and Xu, Minkai and Cao, Ziyao and Lan, Yanyan and Ermon, Stefano and Zhou, Hao and Ma, Wei-Ying},
  journal={Advances in Neural Information Processing Systems},
  volume={36},
  pages={549--568},
  year={2023}
}

@article{lyu2019ultra,
  title={Ultra-large library docking for discovering new chemotypes},
  author={Lyu, Jiankun and Wang, Sheng and Balius, Trent E and Singh, Isha and Levit, Anat and Moroz, Yurii S and O’Meara, Matthew J and Che, Tao and Algaa, Enkhjargal and Tolmachova, Kateryna and others},
  journal={Nature},
  volume={566},
  number={7743},
  pages={224--229},
  year={2019},
  publisher={Nature Publishing Group UK London}
}

@article{zhou2024artificial,
  title={An artificial intelligence accelerated virtual screening platform for drug discovery},
  author={Zhou, Guangfeng and Rusnac, Domnita-Valeria and Park, Hahnbeom and Canzani, Daniele and Nguyen, Hai Minh and Stewart, Lance and Bush, Matthew F and Nguyen, Phuong Tran and Wulff, Heike and Yarov-Yarovoy, Vladimir and others},
  journal={Nature Communications},
  volume={15},
  number={1},
  pages={7761},
  year={2024},
  publisher={Nature Publishing Group UK London}
}

@article{jeon2025stella,
  title={STELLA provides a drug design framework enabling extensive fragment-level chemical space exploration and balanced multi-parameter optimization},
  author={Jeon, Hokyun and Lee, Jin Gyu and Shin, Wonseok and Ji, Hyunjun and Joung, InSuk and Lee, Hui Sun},
  journal={Scientific Reports},
  volume={15},
  number={1},
  pages={28135},
  year={2025},
  publisher={Nature Publishing Group UK London}
}

@article{cai2025fragment,
  title={Fragment-Driven Progressive Alternating Diffusion for De Novo Molecular Design},
  author={Cai, Xing and Zhang, Tong and Qiu, Yide and Cui, Zhen},
  journal={IEEE Transactions on Computational Biology and Bioinformatics},
  year={2025},
  publisher={IEEE}
}

@article{li2024deep,
  title={A deep learning approach for rational ligand generation with toxicity control via reactive building blocks},
  author={Li, Pengyong and Zhang, Kaihao and Liu, Tianxiao and Lu, Ruiqiang and Chen, Yangyang and Yao, Xiaojun and Gao, Lin and Zeng, Xiangxiang},
  journal={Nature Computational Science},
  pages={1--14},
  year={2024},
  publisher={Nature Publishing Group US New York}
}

@article{dpo,
  title={Direct preference optimization: Your language model is secretly a reward model},
  author={Rafailov, Rafael and Sharma, Archit and Mitchell, Eric and Manning, Christopher D and Ermon, Stefano and Finn, Chelsea},
  journal={Advances in Neural Information Processing Systems},
  volume={36},
  pages={53728--53741},
  year={2023}
}

@inproceedings{peng2022pocket2mol,
  title={Pocket2mol: Efficient molecular sampling based on 3d protein pockets},
  author={Peng, Xingang and Luo, Shitong and Guan, Jiaqi and Xie, Qi and Peng, Jian and Ma, Jianzhu},
  booktitle={International Conference on Machine Learning},
  pages={17644--17655},
  year={2022},
  organization={PMLR}
}

@article{safegpt,
  title={Gotta be SAFE: a new framework for molecular design},
  author={Noutahi, Emmanuel and Gabellini, Cristian and Craig, Michael and Lim, Jonathan SC and Tossou, Prudencio},
  journal={Digital Discovery},
  volume={3},
  number={4},
  pages={796--804},
  year={2024},
  publisher={Royal Society of Chemistry}
}

@article{genmol,
  title={GenMol: A Drug Discovery Generalist with Discrete Diffusion},
  author={Lee, Seul and Kreis, Karsten and Veccham, Srimukh Prasad and Liu, Meng and Reidenbach, Danny and Peng, Yuxing and Paliwal, Saee and Nie, Weili and Vahdat, Arash},
  journal={arXiv preprint arXiv:2501.06158},
  year={2025}
}

@inproceedings{geam,
  title={Drug Discovery with Dynamic Goal-aware Fragments},
  author={Lee, Seul and Lee, Seanie and Kawaguchi, Kenji and Hwang, Sung Ju},
  booktitle={International Conference on Machine Learning},
  pages={26731--26751},
  year={2024},
  organization={PMLR}
}

@article{reinvent,
  title={Molecular de-novo design through deep reinforcement learning},
  author={Olivecrona, Marcus and Blaschke, Thomas and Engkvist, Ola and Chen, Hongming},
  journal={Journal of cheminformatics},
  volume={9},
  pages={1--14},
  year={2017},
  publisher={Springer}
}

@article{graphga,
  title={A graph-based genetic algorithm and generative model/Monte Carlo tree search for the exploration of chemical space},
  author={Jensen, Jan H},
  journal={Chemical science},
  volume={10},
  number={12},
  pages={3567--3572},
  year={2019},
  publisher={Royal Society of Chemistry}
}

@article{morld,
  title={Autonomous molecule generation using reinforcement learning and docking to develop potential novel inhibitors},
  author={Jeon, Woosung and Kim, Dongsup},
  journal={Scientific reports},
  volume={10},
  number={1},
  pages={22104},
  year={2020},
  publisher={Nature Publishing Group UK London}
}

@inproceedings{hiervae,
  title={Hierarchical generation of molecular graphs using structural motifs},
  author={Jin, Wengong and Barzilay, Regina and Jaakkola, Tommi},
  booktitle={International conference on machine learning},
  pages={4839--4848},
  year={2020},
  organization={PMLR}
}

@inproceedings{rationale,
  title={Multi-objective molecule generation using interpretable substructures},
  author={Jin, Wengong and Barzilay, Regina and Jaakkola, Tommi},
  booktitle={International conference on machine learning},
  pages={4849--4859},
  year={2020},
  organization={PMLR}
}

@article{freed,
  title={Hit and lead discovery with explorative rl and fragment-based molecule generation},
  author={Yang, Soojung and Hwang, Doyeong and Lee, Seul and Ryu, Seongok and Hwang, Sung Ju},
  journal={Advances in Neural Information Processing Systems},
  volume={34},
  pages={7924--7936},
  year={2021}
}

@inproceedings{mood,
  title={Conditional graph information bottleneck for molecular relational learning},
  author={Lee, Namkyeong and Hyun, Dongmin and Na, Gyoung S and Kim, Sungwon and Lee, Junseok and Park, Chanyoung},
  booktitle={International Conference on Machine Learning},
  pages={18852--18871},
  year={2023},
  organization={PMLR}
}

@article{f-rag,
  title={Molecule generation with fragment retrieval augmentation},
  author={Lee, Seul and Kreis, Karsten and Veccham, Srimukh and Liu, Meng and Reidenbach, Danny and Paliwal, Saee and Vahdat, Arash and Nie, Weili},
  journal={Advances in Neural Information Processing Systems},
  volume={37},
  pages={132463--132490},
  year={2024}
}

@inproceedings{jtvae,
  title={Junction tree variational autoencoder for molecular graph generation},
  author={Jin, Wengong and Barzilay, Regina and Jaakkola, Tommi},
  booktitle={International conference on machine learning},
  pages={2323--2332},
  year={2018},
  organization={PMLR}
}

@article{mars,
  title={Mars: Markov molecular sampling for multi-objective drug discovery},
  author={Xie, Yutong and Shi, Chence and Zhou, Hao and Yang, Yuwei and Zhang, Weinan and Yu, Yong and Li, Lei},
  journal={arXiv preprint arXiv:2103.10432},
  year={2021}
}

@article{pavae,
  title={Molecule generation by principal subgraph mining and assembling},
  author={Kong, Xiangzhe and Huang, Wenbing and Tan, Zhixing and Liu, Yang},
  journal={Advances in Neural Information Processing Systems},
  volume={35},
  pages={2550--2563},
  year={2022}
}

@article{gegl,
  title={Guiding deep molecular optimization with genetic exploration},
  author={Ahn, Sungsoo and Kim, Junsu and Lee, Hankook and Shin, Jinwoo},
  journal={Advances in neural information processing systems},
  volume={33},
  pages={12008--12021},
  year={2020}
}

@inproceedings{gflow,
  title={Genetic-guided GFlowNets for Sample Efficient Molecular Optimization},
  author={Kim, Hyeonah and Kim, Minsu and Choi, Sanghyeok and Park, Jinkyoo},
  booktitle={The Thirty-eighth Annual Conference on Neural Information Processing Systems},
  year={2024}
}

@article{g+d,
  title={Augmenting genetic algorithms with deep neural networks for exploring the chemical space},
  author={Nigam, AkshatKumar and Friederich, Pascal and Krenn, Mario and Aspuru-Guzik, Al{\'a}n},
  journal={arXiv preprint arXiv:1909.11655},
  year={2019}
}

@inproceedings{retmol,
  title={Retrieval-based Controllable Molecule Generation},
  author={Wang, Zichao and Nie, Weili and Qiao, Zhuoran and Xiao, Chaowei and Baraniuk, Richard and Anandkumar, Anima},
  booktitle={International Conference on Learning Representations (ICLR) 2023},
  year={2023}
}

@article{schneuing2024structure,
  title={Structure-based drug design with equivariant diffusion models},
  author={Schneuing, Arne and Harris, Charles and Du, Yuanqi and Didi, Kieran and Jamasb, Arian and Igashov, Ilia and Du, Weitao and Gomes, Carla and Blundell, Tom L and Lio, Pietro and others},
  journal={Nature Computational Science},
  volume={4},
  number={12},
  pages={899--909},
  year={2024},
  publisher={Nature Publishing Group US New York}
}

@article{lin2025diffbp,
  title={Diffbp: Generative diffusion of 3d molecules for target protein binding},
  author={Lin, Haitao and Huang, Yufei and Zhang, Odin and Ma, Siqi and Liu, Meng and Li, Xuanjing and Wu, Lirong and Wang, Jishui and Hou, Tingjun and Li, Stan Z},
  journal={Chemical Science},
  volume={16},
  number={3},
  pages={1417--1431},
  year={2025},
  publisher={Royal Society of Chemistry}
}

@article{zhang2023resgen,
  title={ResGen is a pocket-aware 3D molecular generation model based on parallel multiscale modelling},
  author={Zhang, Odin and Zhang, Jintu and Jin, Jieyu and Zhang, Xujun and Hu, RenLing and Shen, Chao and Cao, Hanqun and Du, Hongyan and Kang, Yu and Deng, Yafeng and others},
  journal={Nature Machine Intelligence},
  volume={5},
  number={9},
  pages={1020--1030},
  year={2023},
  publisher={Nature Publishing Group UK London}
}

@article{zhang2024fraggen,
  title={FragGen: towards 3D geometry reliable fragment-based molecular generation},
  author={Zhang, Odin and Huang, Yufei and Cheng, Shichen and Yu, Mengyao and Zhang, Xujun and Lin, Haitao and Zeng, Yundian and Wang, Mingyang and Wu, Zhenxing and Zhao, Huifeng and others},
  journal={Chemical Science},
  volume={15},
  number={46},
  pages={19452--19465},
  year={2024},
  publisher={Royal Society of Chemistry}
}

@article{guan20233d,
  title={3d equivariant diffusion for target-aware molecule generation and affinity prediction},
  author={Guan, Jiaqi and Qian, Wesley Wei and Peng, Xingang and Su, Yufeng and Peng, Jian and Ma, Jianzhu},
  journal={arXiv preprint arXiv:2303.03543},
  year={2023}
}

@inproceedings{podda2020deep,
  title={A deep generative model for fragment-based molecule generation},
  author={Podda, Marco and Bacciu, Davide and Micheli, Alessio},
  booktitle={International conference on artificial intelligence and statistics},
  pages={2240--2250},
  year={2020},
  organization={PMLR}
}

@article{degen2008art,
  title={On the art of compiling and using'drug-like'chemical fragment spaces},
  author={Degen, Jorg and Wegscheid-Gerlach, Christof and Zaliani, Andrea and Rarey, Matthias},
  journal={ChemMedChem},
  volume={3},
  number={10},
  pages={1503},
  year={2008}
}

@inproceedings{xie2023much,
  title={HOW MUCH SPACE HAS BEEN EXPLORED? MEASURING THE CHEMICAL SPACE COVERED BY DATABASES AND MACHINE-GENERATED MOLECULES},
  author={Xie, Yutong and Xu, Ziqiao and Ma, Jiaqi and Mei, Qiaozhu},
  booktitle={11th International Conference on Learning Representations, ICLR 2023},
  year={2023}
}

@article{schwaller2019molecular,
  title={Molecular transformer: a model for uncertainty-calibrated chemical reaction prediction},
  author={Schwaller, Philippe and Laino, Teodoro and Gaudin, Th{\'e}ophile and Bolgar, Peter and Hunter, Christopher A and Bekas, Costas and Lee, Alpha A},
  journal={ACS central science},
  volume={5},
  number={9},
  pages={1572--1583},
  year={2019},
  publisher={ACS Publications}
}

@article{su2024roformer,
  title={Roformer: Enhanced transformer with rotary position embedding},
  author={Su, Jianlin and Ahmed, Murtadha and Lu, Yu and Pan, Shengfeng and Bo, Wen and Liu, Yunfeng},
  journal={Neurocomputing},
  volume={568},
  pages={127063},
  year={2024},
  publisher={Elsevier}
}

@article{olivecrona2017molecular,
  title={Molecular de-novo design through deep reinforcement learning},
  author={Olivecrona, Marcus and Blaschke, Thomas and Engkvist, Ola and Chen, Hongming},
  journal={Journal of cheminformatics},
  volume={9},
  pages={1--14},
  year={2017},
  publisher={Springer}
}

@inproceedings{chaslot2008monte,
  title={Monte-carlo tree search: A new framework for game ai},
  author={Chaslot, Guillaume and Bakkes, Sander and Szita, Istvan and Spronck, Pieter},
  booktitle={Proceedings of the AAAI Conference on Artificial Intelligence and Interactive Digital Entertainment},
  volume={4},
  number={1},
  pages={216--217},
  year={2008}
}

@article{auer2002finite,
  title={Finite-time analysis of the multiarmed bandit problem},
  author={Auer, Peter and Cesa-Bianchi, Nicolo and Fischer, Paul},
  journal={Machine learning},
  volume={47},
  pages={235--256},
  year={2002},
  publisher={Springer}
}

@article{tian2024toward,
  title={Toward self-improvement of llms via imagination, searching, and criticizing},
  author={Tian, Ye and Peng, Baolin and Song, Linfeng and Jin, Lifeng and Yu, Dian and Han, Lei and Mi, Haitao and Yu, Dong},
  journal={Advances in Neural Information Processing Systems},
  volume={37},
  pages={52723--52748},
  year={2024}
}

@article{ppo,
  title={Proximal policy optimization algorithms},
  author={Schulman, John and Wolski, Filip and Dhariwal, Prafulla and Radford, Alec and Klimov, Oleg},
  journal={arXiv preprint arXiv:1707.06347},
  year={2017}
}

@article{zinc,
  title={ZINC- a free database of commercially available compounds for virtual screening},
  author={Irwin, John J and Shoichet, Brian K},
  journal={Journal of chemical information and modeling},
  volume={45},
  number={1},
  pages={177--182},
  year={2005},
  publisher={ACS Publications}
}

@article{unichem,
  title={UniChem: a unified chemical structure cross-referencing and identifier tracking system},
  author={Chambers, Jon and Davies, Mark and Gaulton, Anna and Hersey, Anne and Velankar, Sameer and Petryszak, Robert and Hastings, Janna and Bellis, Louisa and McGlinchey, Shaun and Overington, John P},
  journal={Journal of cheminformatics},
  volume={5},
  number={1},
  pages={3},
  year={2013},
  publisher={Springer}
}
\bibliographystyle{naturemag}

% *** END NOTES ***
% \newpage
\subsection*{Acknowledgements}
This work was supported by the National Natural Science Foundation of China under Grant 6247617, and the Guangdong Natural Science Foundation Project under Grant 2025A1515011567.

\subsection*{Author Contributions}
Z.Y. and J.J. conceived the idea for Trio. J.J., R.B., and Z.Z. coordinated and supervised the project. Z.Y. designed and implemented the complete workflow and algorithms of Trio, with contributions from D.X. to the implementation. Z.Y. and D.X. conducted the experiments, analyzed the data, and drafted the manuscript. J.J., J.L., R.B., T.H., and Z.Z. provided critical feedback on the algorithm evaluation and revised the manuscript.

\subsection*{Declaration of Interests}
The authors declare no competing interests.

\end{document}